\documentclass{article}

\usepackage{microtype}
\usepackage{graphicx}
\usepackage{booktabs} 

\usepackage{hyperref}

\usepackage[accepted]{xxxxxx2025}

\usepackage{amsmath}
\usepackage{amssymb}
\usepackage{mathtools}
\usepackage{amsthm}

\usepackage[capitalize,noabbrev]{cleveref}

\usepackage{pifont}         
\usepackage{tikz}
\usetikzlibrary{shapes.geometric}
\usepackage{amsthm, amssymb}
\usepackage{bm}
\usepackage{caption}
\usepackage{subcaption} 
\usepackage{rotating} 

\theoremstyle{plain}
\newtheorem{theorem}{Theorem}[section]

\theoremstyle{definition}
\newtheorem{definition}[theorem]{Definition}

\theoremstyle{remark}

\newtheorem{proofsketch}{Proof (Step-by-Step)}

\usepackage[textsize=small]{todonotes}

\xxxxxxtitlerunning{Privilege Scores}

\begin{document}
\ifdefined\N                                                                
\renewcommand{\N}{\mathds{N}} 
\else \newcommand{\N}{\mathds{N}} \fi 
\newcommand{\Z}{\mathds{Z}} 
\newcommand{\Q}{\mathds{Q}} 
\newcommand{\R}{\mathds{R}} 
\ifdefined\C 
  \renewcommand{\C}{\mathds{C}} 
\else \newcommand{\C}{\mathds{C}} \fi
\newcommand{\continuous}{\mathcal{C}} 
\newcommand{\M}{\mathcal{M}} 
\newcommand{\epsm}{\epsilon_m} 

\newcommand{\setzo}{\{0, 1\}} 
\newcommand{\setmp}{\{-1, +1\}} 
\newcommand{\unitint}{[0, 1]} 

\newcommand{\xt}{\tilde x} 
\newcommand{\argmax}{\operatorname{arg\,max}} 
\newcommand{\argmin}{\operatorname{arg\,min}} 
\newcommand{\argminlim}{\mathop{\mathrm{arg\,min}}\limits} 
\newcommand{\argmaxlim}{\mathop{\mathrm{arg\,max}}\limits} 
\newcommand{\sign}{\operatorname{sign}} 
\newcommand{\I}{\mathbb{I}} 
\newcommand{\order}{\mathcal{O}} 
\newcommand{\pd}[2]{\frac{\partial{#1}}{\partial #2}} 
\newcommand{\floorlr}[1]{\left\lfloor #1 \right\rfloor} 
\newcommand{\ceillr}[1]{\left\lceil #1 \right\rceil} 

\newcommand{\sumin}{\sum\limits_{i=1}^n} 
\newcommand{\sumim}{\sum\limits_{i=1}^m} 
\newcommand{\sumjn}{\sum\limits_{j=1}^n} 
\newcommand{\sumjp}{\sum\limits_{j=1}^p} 
\newcommand{\sumik}{\sum\limits_{i=1}^k} 
\newcommand{\sumkg}{\sum\limits_{k=1}^g} 
\newcommand{\sumjg}{\sum\limits_{j=1}^g} 
\newcommand{\meanin}{\frac{1}{n} \sum\limits_{i=1}^n} 
\newcommand{\meanim}{\frac{1}{m} \sum\limits_{i=1}^m} 
\newcommand{\meankg}{\frac{1}{g} \sum\limits_{k=1}^g} 
\newcommand{\prodin}{\prod\limits_{i=1}^n} 
\newcommand{\prodkg}{\prod\limits_{k=1}^g} 
\newcommand{\prodjp}{\prod\limits_{j=1}^p} 

\newcommand{\one}{\boldsymbol{1}} 
\newcommand{\zero}{\mathbf{0}} 
\newcommand{\id}{\boldsymbol{I}} 
\newcommand{\diag}{\operatorname{diag}} 
\newcommand{\trace}{\operatorname{tr}} 
\newcommand{\spn}{\operatorname{span}} 
\newcommand{\scp}[2]{\left\langle #1, #2 \right\rangle} 
\newcommand{\mat}[1]{\begin{pmatrix} #1 \end{pmatrix}} 
\newcommand{\Amat}{\mathbf{A}} 
\newcommand{\Deltab}{\mathbf{\Delta}} 

\renewcommand{\P}{\mathbb{P}} 
\newcommand{\E}{\mathbb{E}} 
\newcommand{\var}{\mathsf{Var}} 
\newcommand{\cov}{\mathsf{Cov}} 
\newcommand{\corr}{\mathsf{Corr}} 
\newcommand{\normal}{\mathcal{N}} 
\newcommand{\iid}{\overset{i.i.d}{\sim}} 
\newcommand{\distas}[1]{\overset{#1}{\sim}} 

\newcommand{\Xspace}{\mathcal{X}} 
\newcommand{\Yspace}{\mathcal{Y}} 
\newcommand{\nset}{\{1, \ldots, n\}} 
\newcommand{\pset}{\{1, \ldots, p\}} 
\newcommand{\gset}{\{1, \ldots, g\}} 
\newcommand{\Pxy}{\mathbb{P}_{xy}} 
\newcommand{\Exy}{\mathbb{E}_{xy}} 
\newcommand{\xv}{\mathbf{x}} 
\newcommand{\xtil}{\tilde{\mathbf{x}}} 
\newcommand{\yv}{\mathbf{y}} 
\newcommand{\xy}{(\xv, y)} 
\newcommand{\xvec}{\left(x_1, \ldots, x_p\right)^T} 
\newcommand{\Xmat}{\mathbf{X}} 
\newcommand{\allDatasets}{\mathds{D}} 
\newcommand{\allDatasetsn}{\mathds{D}_n}  
\newcommand{\D}{\mathcal{D}} 
\newcommand{\Dn}{\D_n} 
\newcommand{\Dtrain}{\mathcal{D}_{\text{train}}} 
\newcommand{\Dtest}{\mathcal{D}_{\text{test}}} 
\newcommand{\xyi}[1][i]{\left(\xv^{(#1)}, y^{(#1)}\right)} 
\newcommand{\Dset}{\left( \xyi[1], \ldots, \xyi[n]\right)} 
\newcommand{\defAllDatasetsn}{(\Xspace \times \Yspace)^n} 
\newcommand{\defAllDatasets}{\bigcup_{n \in \N}(\Xspace \times \Yspace)^n} 
\newcommand{\xdat}{\left\{ \xv^{(1)}, \ldots, \xv^{(n)}\right\}} 
\newcommand{\yvec}{\left(y^{(1)}, \hdots, y^{(n)}\right)^T} 
\renewcommand{\xi}[1][i]{\xv^{(#1)}} 
\newcommand{\yi}[1][i]{y^{(#1)}} 
\newcommand{\xivec}{\left(x^{(i)}_1, \ldots, x^{(i)}_p\right)^T} 
\newcommand{\xj}{\xv_j} 
\newcommand{\xjvec}{\left(x^{(1)}_j, \ldots, x^{(n)}_j\right)^T} 
\newcommand{\phiv}{\mathbf{\phi}} 
\newcommand{\phixi}{\mathbf{\phi}^{(i)}} 

\newcommand{\lamv}{\bm{\lambda}} 
\newcommand{\Lam}{\bm{\Lambda}}	 
\newcommand{\preimageInducer}{\left(\defAllDatasets\right)\times\Lam} 
\newcommand{\preimageInducerShort}{\allDatasets\times\Lam} 
\newcommand{\ind}{\mathcal{I}} 

\newcommand{\ftrue}{f_{\text{true}}}  
\newcommand{\ftruex}{\ftrue(\xv)} 
\newcommand{\fx}{f(\xv)} 
\newcommand{\fdomains}{f: \Xspace \rightarrow \R^g} 
\newcommand{\Hspace}{\mathcal{H}} 
\newcommand{\fbayes}{f^{\ast}} 
\newcommand{\fxbayes}{f^{\ast}(\xv)} 
\newcommand{\fkx}[1][k]{f_{#1}(\xv)} 
\newcommand{\fh}{\hat{f}} 
\newcommand{\fxh}{\fh(\xv)} 
\newcommand{\fxt}{f(\xv ~|~ \thetab)} 
\newcommand{\fxi}{f\left(\xv^{(i)}\right)} 
\newcommand{\fxih}{\hat{f}\left(\xv^{(i)}\right)} 
\newcommand{\fxit}{f\left(\xv^{(i)} ~|~ \thetab\right)} 
\newcommand{\fhD}{\fh_{\D}} 
\newcommand{\fhDtrain}{\fh_{\Dtrain}} 
\newcommand{\fhDnlam}{\fh_{\Dn, \lamv}} 
\newcommand{\fhDlam}{\fh_{\D, \lamv}} 
\newcommand{\fhDnlams}{\fh_{\Dn, \lamv^\ast}} 
\newcommand{\fhDlams}{\fh_{\D, \lamv^\ast}} 

\newcommand{\hx}{h(\xv)} 
\newcommand{\hh}{\hat{h}} 
\newcommand{\hxh}{\hat{h}(\xv)} 
\newcommand{\hxt}{h(\xv | \thetab)} 
\newcommand{\hxi}{h\left(\xi\right)} 
\newcommand{\hxit}{h\left(\xi ~|~ \thetab\right)} 
\newcommand{\hbayes}{h^{\ast}} 
\newcommand{\hxbayes}{h^{\ast}(\xv)} 

\newcommand{\yh}{\hat{y}} 
\newcommand{\yih}{\hat{y}^{(i)}} 
\newcommand{\resi}{\yi- \yih}

\newcommand{\thetah}{\hat{\theta}} 
\newcommand{\thetab}{\bm{\theta}} 
\newcommand{\thetabh}{\bm{\hat\theta}} 
\newcommand{\thetat}[1][t]{\thetab^{[#1]}} 
\newcommand{\thetatn}[1][t]{\thetab^{[#1 +1]}} 
\newcommand{\thetahDnlam}{\thetabh_{\Dn, \lamv}} 
\newcommand{\thetahDlam}{\thetabh_{\D, \lamv}} 
\newcommand{\mint}{\min_{\thetab \in \Theta}} 
\newcommand{\argmint}{\argmin_{\thetab \in \Theta}} 

\newcommand{\pdf}{p} 
\newcommand{\pdfx}{p(\xv)} 
\newcommand{\pixt}{\pi(\xv~|~ \thetab)} 
\newcommand{\pixit}{\pi\left(\xi ~|~ \thetab\right)} 
\newcommand{\pixii}{\pi(\xi)} 
\newcommand{\pii}{\pi^{(i)}} 

\newcommand{\pdfxy}{p(\xv,y)} 
\newcommand{\pdfxyt}{p(\xv, y ~|~ \thetab)} 
\newcommand{\pdfxyit}{p\left(\xi, \yi ~|~ \thetab\right)} 

\newcommand{\pdfxyk}[1][k]{p(\xv | y= #1)} 
\newcommand{\lpdfxyk}[1][k]{\log p(\xv | y= #1)} 
\newcommand{\pdfxiyk}[1][k]{p\left(\xi | y= #1 \right)} 

\newcommand{\pik}[1][k]{\pi_{#1}} 
\newcommand{\lpik}[1][k]{\log \pi_{#1}} 
\newcommand{\pit}{\pi(\thetab)} 

\newcommand{\post}{\P(y = 1 ~|~ \xv)} 
\newcommand{\postk}[1][k]{\P(y = #1 ~|~ \xv)} 
\newcommand{\pidomains}{\pi: \Xspace \rightarrow \unitint} 
\newcommand{\pibayes}{\pi^{\ast}} 
\newcommand{\pixbayes}{\pi^{\ast}(\xv)} 
\newcommand{\pix}{\pi(\xv)} 
\newcommand{\pikx}[1][k]{\pi_{#1}(\xv)} 
\newcommand{\pikxt}[1][k]{\pi_{#1}(\xv ~|~ \thetab)} 
\newcommand{\pixh}{\hat \pi(\xv)} 
\newcommand{\pikxh}[1][k]{\hat \pi_{#1}(\xv)} 
\newcommand{\pixih}{\hat \pi(\xi)} 
\newcommand{\pikxih}[1][k]{\hat \pi_{#1}(\xi)} 
\newcommand{\pdfygxt}{p(y ~|~\xv, \thetab)} 
\newcommand{\pdfyigxit}{p\left(\yi ~|~\xi, \thetab\right)} 
\newcommand{\lpdfygxt}{\log \pdfygxt } 
\newcommand{\lpdfyigxit}{\log \pdfyigxit} 

\newcommand{\bayesrulek}[1][k]{\frac{\P(\xv | y= #1) \P(y= #1)}{\P(\xv)}} 
\newcommand{\muk}{\bm{\mu_k}} 

\newcommand{\eps}{\epsilon} 
\newcommand{\epsi}{\epsilon^{(i)}} 
\newcommand{\epsh}{\hat{\epsilon}} 
\newcommand{\yf}{y \fx} 
\newcommand{\yfi}{\yi \fxi} 
\newcommand{\Sigmah}{\hat \Sigma} 
\newcommand{\Sigmahj}{\hat \Sigma_j} 

\newcommand{\Lyf}{L\left(y, f\right)} 
\newcommand{\Lxy}{L\left(y, \fx\right)} 
\newcommand{\Lxyi}{L\left(\yi, \fxi\right)} 
\newcommand{\Lxyt}{L\left(y, \fxt\right)} 
\newcommand{\Lxyit}{L\left(\yi, \fxit\right)} 
\newcommand{\Lxym}{L\left(\yi, f\left(\bm{\tilde{x}}^{(i)} ~|~ \thetab\right)\right)} 
\newcommand{\Lpixy}{L\left(y, \pix\right)} 
\newcommand{\Lpixyi}{L\left(\yi, \pixii\right)} 
\newcommand{\Lpixyt}{L\left(y, \pixt\right)} 
\newcommand{\Lpixyit}{L\left(\yi, \pixit\right)} 
\newcommand{\Lhxy}{L\left(y, \hx\right)} 
\newcommand{\Lr}{L\left(r\right)} 
\newcommand{\lone}{|y - \fx|} 
\newcommand{\ltwo}{\left(y - \fx\right)^2} 
\newcommand{\lbernoullimp}{\ln(1 + \exp(-y \cdot \fx))} 
\newcommand{\lbernoullizo}{- y \cdot \fx + \log(1 + \exp(\fx))} 
\newcommand{\lcrossent}{- y \log \left(\pix\right) - (1 - y) \log \left(1 - \pix\right)} 
\newcommand{\lbrier}{\left(\pix - y \right)^2} 
\newcommand{\risk}{\mathcal{R}} 
\newcommand{\riskbayes}{\mathcal{R}^\ast}
\newcommand{\riskf}{\risk(f)} 
\newcommand{\riskdef}{\E_{y|\xv}\left(\Lxy \right)} 
\newcommand{\riskt}{\mathcal{R}(\thetab)} 
\newcommand{\riske}{\mathcal{R}_{\text{emp}}} 
\newcommand{\riskeb}{\bar{\mathcal{R}}_{\text{emp}}} 
\newcommand{\riskef}{\riske(f)} 
\newcommand{\risket}{\mathcal{R}_{\text{emp}}(\thetab)} 
\newcommand{\riskr}{\mathcal{R}_{\text{reg}}} 
\newcommand{\riskrt}{\mathcal{R}_{\text{reg}}(\thetab)} 
\newcommand{\riskrf}{\riskr(f)} 
\newcommand{\riskrth}{\hat{\mathcal{R}}_{\text{reg}}(\thetab)} 
\newcommand{\risketh}{\hat{\mathcal{R}}_{\text{emp}}(\thetab)} 
\newcommand{\LL}{\mathcal{L}} 
\newcommand{\LLt}{\mathcal{L}(\thetab)} 
\newcommand{\LLtx}{\mathcal{L}(\thetab | \xv)} 
\newcommand{\logl}{\ell} 
\newcommand{\loglt}{\logl(\thetab)} 
\newcommand{\logltx}{\logl(\thetab | \xv)} 
\newcommand{\errtrain}{\text{err}_{\text{train}}} 
\newcommand{\errtest}{\text{err}_{\text{test}}} 
\newcommand{\errexp}{\overline{\text{err}_{\text{test}}}} 

\newcommand{\thx}{\thetab^T \xv} 
\newcommand{\olsest}{(\Xmat^T \Xmat)^{-1} \Xmat^T \yv} 

\newcommand{\topic}[1]{\textbf{[#1]}}

\newcommand{\xti}{\xtil^{(i)}}
\newcommand{\yti}{\tilde{y}^{(i)}}
\newcommand{\yt}{\tilde{y}}

\newcommand{\pitil}{\varphi}
\newcommand{\pitilh}{\hat{\pitil}}
\newcommand{\pitilhf}{\pitilh(\cdot)}
\newcommand{\piti}{\pitil^{(i)}}
\newcommand{\pitj}{\pitil^{(j)}}
\newcommand{\pitifh}{\pitilh(\cdot)}
\newcommand{\pitxtih}{\pitilh(\xti)} 
\newcommand{\pitxth}{\pitilh(\xtil)} 
\newcommand{\pitxih}{\pitilh(\xi)} 
\newcommand{\pitxti}{\pitil(\xti)}
\newcommand{\pitxi}{\pitil(\xi)}
\newcommand{\pixv}{\pi(\xv)}

\newcommand{\pitx}{\pitil(\xtil)}
\newcommand{\psix}{\psi(\xv_F)}

\newcommand{\pitxh}{\pitilh(\xtil)}
\newcommand{\psixh}{\psih(\xv_F)}

\newcommand{\spiti}{s(\piti)}
\newcommand{\spitxtih}{s(\pitxtih)}
\newcommand{\spitxih}{s(\pitxih)}
\newcommand{\spitxti}{s(\pitxti)}
\newcommand{\spitxi}{s(\pitxi)}

\newcommand{\spii}{s(\pii)}
\newcommand{\spixih}{s(\pixih)}
\newcommand{\spixii}{s(\pixii)}

\newcommand{\psih}{\hat{\psi}}
\newcommand{\psihi}{\psih^{(i)}}
\newcommand{\psii}{\psi^{(i)}}
\newcommand{\psij}{\psi^{(j)}}

\newcommand{\psif}{\psi(\cdot)}
\newcommand{\psihf}{\hat{\psi}(\cdot)}

\newcommand{\spsii}{s(\psii)}
\newcommand{\psixi}{\psi(\xfindi)}
\newcommand{\psixih}{\psih(\xfindi)}
\newcommand{\spsixii}{s(\psixi)}
\newcommand{\spsixih}{s(\psixih)}

\newcommand{\ti}{t^{(i)}}
\newcommand{\tj}{t^{(j)}}
\newcommand{\wi}{m^{(i)}}
\newcommand{\w}{m}
\newcommand{\W}{M}
\newcommand{\wj}{m^{(j)}}
\newcommand{\zi}{z^{(i)}}
\newcommand{\vi}{v^{(i)}}
\newcommand{\mui}{\mu^{(i)}}

\newcommand{\sfu}{s(\cdot)}
\newcommand{\pij}{\pi^{(j)}}
\newcommand{\pif}{\pi(\cdot)}
\newcommand{\pih}{\hat{\pi}}
\newcommand{\pihi}{\pih^{(i)}}
\newcommand{\pihj}{\pih^{(j)}}
\newcommand{\pihf}{\pih(\cdot)}
\newcommand{\pitf}{\pitil(\cdot)}
\newcommand{\pith}{\hat{\pitil}}
\newcommand{\pithf}{\hat{\pitil}(\cdot)}
\newcommand{\pithi}{\hat{\pitil}^{(i)}}
\newcommand{\pithj}{\hat{\pitil}^{(j)}}

\newcommand{\xyti}[1][i]{\left(\xtil^{(#1)}, \yt^{(#1)}\right)} 
\newcommand{\Dsett}{\left( \xyti[1], \ldots, \xyti[n]\right)} 
\newcommand{\Dt}{\tilde{\mathcal{D}}} 

\newcommand{\Xspacet}{\tilde{\Xspace}}

\newcommand{\xfindi}{\xv_F^{(i)}}
\newcommand{\xfindj}{\xv_F^{(j)}}
\newcommand{\yfindi}{y_F^{(i)}}
\newcommand{\yfind}{y_F}

\newcommand{\priv}{\delta}
\newcommand{\privi}{\priv^{(i)}}
\newcommand{\prob}{\mathbb{P}}
\newcommand{\hp}{\hat{\priv}}
\newcommand{\hpi}{\hp^{(i)}}
\newcommand{\hpx}[1][]{\hp\if\relax\detokenize{#1}\relax(\xv)\else_{#1}(\xv)\fi}
\newcommand{\px}[1][]{\priv\if\relax\detokenize{#1}\relax(\xv)\else_{#1}(\xv)\fi}
\newcommand{\hpxtil}[1][]{\hp\if\relax\detokenize{#1}\relax(\xtil)\else_{#1}(\xtil)\fi}
\newcommand{\pxtil}[1][]{\priv\if\relax\detokenize{#1}\relax(\xtil)\else_{#1}(\xtil)\fi}
\newcommand{\tgx}[1][]{\tilde{\gamma}_{#1}(\xv)}
\newcommand{\gx}[1][]{\gamma_{#1}(\xv)}
\newcommand{\hgx}[1][]{\hat{\gamma}_{#1}(\xv)}

\newcommand{\ai}{\mathbf{a}^{(i)}}

\newcommand{\cih}{\widehat{CI}}
\newcommand{\cihi}{\widehat{CI}^{(i)}}
\newcommand{\cihia}{\cihi_{\alpha}}

\newcommand{\lb}[1]{\textcolor{blue}{[L: #1]}}
\newcommand{\ja}[1]{\textcolor{orange}{[J: #1]}}
\newcommand{\kp}[1]{\textcolor{blue}{[K: #1]}}
\newcommand{\bp}[1]{\textcolor{blue}{$\rightarrow$ #1 \\}}

\newcommand{\myemph}[1]{\textit{#1}}

\newcommand{\mydef}[1]{\paragraph{#1}}

\twocolumn[
\xxxxxxtitle{Privilege Scores}



\xxxxxxsetsymbol{equal}{*}

\begin{xxxxxxauthorlist}
\xxxxxxauthor{Ludwig Bothmann}{yyy,xxx}
\xxxxxxauthor{Philip A. Boustani}{yyy,xxx}
\xxxxxxauthor{Jose M. Alvarez}{zzz}
\xxxxxxauthor{Giuseppe Casalicchio}{yyy,xxx}
\xxxxxxauthor{Bernd Bischl}{yyy,xxx}
\xxxxxxauthor{Susanne Dandl}{aaa}
\end{xxxxxxauthorlist}

\xxxxxxaffiliation{yyy}{Department of Statistics, LMU Munich, Germany}
\xxxxxxaffiliation{xxx}{Munich Center for Machine Learning (MCML), Munich, Germany}
\xxxxxxaffiliation{zzz}{Department of Computer Science, KU Leuven, Belgium}
\xxxxxxaffiliation{aaa}{Epidemiology, Biostatistics \& Prevention Institute  (EBPI), Universität Zürich, Zurich, Switzerland}

\xxxxxxcorrespondingauthor{Ludwig Bothmann}{ludwig.bothmann@lmu.de}

\xxxxxxkeywords{Machine Learning, xxxxxx}

\vskip 0.3in
]



\printAffiliationsAndNotice{}  

\begin{abstract}
Bias-transforming methods of fairness-aware machine learning aim to correct a non-neutral status quo with respect to a protected attribute (PA).
Current methods, however, lack an explicit formulation of what drives non-neutrality. 
We introduce \textit{privilege scores (PS)} to measure PA-related privilege by comparing the model predictions in the real world with those in a fair world in which the influence of the PA is removed.
At the individual level, PS can identify individuals who qualify for affirmative action;
at the global level, PS can inform bias-transforming policies.
After presenting estimation methods for PS, we propose \textit{privilege score contributions (PSCs)}, an interpretation method that attributes the origin of privilege to mediating features and direct effects. 
We provide confidence intervals for both PS and PSCs.
Experiments on simulated and real-world data demonstrate the broad applicability of our methods and provide novel insights into gender and racial privilege in mortgage and college admissions applications.
%
\end{abstract}

\section{Introduction}


Fairness-aware machine learning (fairML) 
methods can be classified into ``bias-preserving'' and ``bias-transforming'' approaches.
The former assume a \textit{neutral status quo} (referring to absence of real-world discrimination) and pursue \textit{formal equality} by balancing model error rates across subgroups based on membership to the protected attribute (PA). 
Such methods ``aim to not make society more unequal than the status quo''.
The latter, instead, assume a \textit{non-neutral status quo} and pursue \textit{substantive equality} by correcting the systematic biases behind the model error rates.
Such methods account for ``historical inequalities which actively ought to be eroded'' \cite{wachter_bias_2021}.

\begin{figure}[ht]
    \centering
    \includegraphics[width=0.5\textwidth]{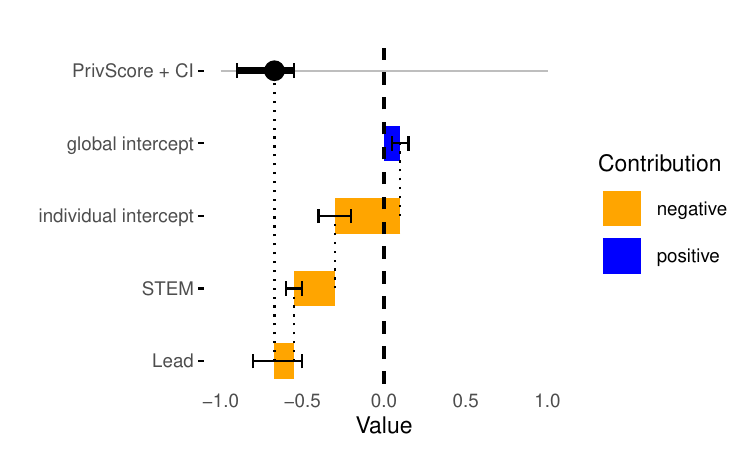} 
    \caption{PS and PS contributions for Amina.}
    \label{fig:iml_viz}
\end{figure}

Rectifying a non-neutral status quo is increasingly studied in fairML \citep[e.g.,][]{Alvarez2024Policy, Mittelstadt2024UnfairML, Russo2024BridgingResearchAndPractice}.
Motivated by discussions on non-discrimination law \citep[e.g.,][]{issa_kohler-hausmann_eddie_2019, wachter_bias_2021, weerts_algorithmic_2023}, including affirmative action \cite{Romei2014MultiSurveyDiscrimination}, a growing body of fairML work targets substantive equality and proposes bias-transforming methods  \citep[e.g.,][]{kusner_counterfactual_2017, black_fliptest_2020, plecko_fair_2020, alvarez_counterfactual_2023, bothmann_causal_2023}.
With varying degrees of explicitness, these works account for the influence of the PA on the other attributes to ``imagine'' an alternative and fairer world, particularly at the level of individual representation.
There is, however, a lack of articulation 
on what exactly is being rectified.
Rectifying systematic biases and, ultimately, discriminatory patterns, indeed motivate these works, but these notions can be vague 
beyond the legal realm.
From a machine learning (ML) point of view, a more explicit formulation is needed for what we measure and wish to address when implementing bias-transforming fairML.

This paper introduces \textit{privilege scores (PS)}.
Assuming a non-neutral status quo, PS measure the level of privilege 
due to the PA experienced by individuals within a decision-making context. 
We compute the PS by comparing the output of ML models between two worlds: the real world and a ``fair world'' in which the PA has no causal effect on the target.
At the individual level, PS identify 
those individuals worthy of bias-transforming methods.
At the global level, PS not only allow to quantify 
group privileges, but also explain the mediators through which these privileges are manifested, making PS a method for informing bias-transforming policies.
%
%
The proposed PS apply at the dataset, model, and decision stage levels.
In addition to point estimates of PS, we provide individual confidence intervals that allow us to judge whether privilege is significantly non-zero. 
We also develop an interpretation method based on Shapley values for PS to quantify each feature's contribution to an individual's privilege.


\paragraph{Main contributions.} 
(1) We propose the formal concept of \textit{privilege scores} (PS) to quantify the impact of PAs on decision-making outcomes by comparing real-world scenarios to a ``fair world'' baseline; 
(2) we propose a methodological framework for (i) estimating PS, and (ii) quantifying the uncertainty of the estimators; 
(3) we present \textit{privilege score contributions (PSCs)}, 
an interpretation method to quantify each feature's contribution to an individual's privilege; 
(4) we perform extensive experiments to showcase the performance of PS on simulated and real-world data.

\subsection{Motivating examples}
\label{sec:usecases}

The following fictional examples motivate the use for PS at three levels: decision stage (individual assessment and affirmative action; see Appendix \ref{app:sec:uc_aff}), model (auditing ADM), and data set (auditing real-world DGP).
\paragraph{Individual assessment} 
\label{sec:uc_ind}

Amina applies for a mortgage. The bank's automated decision making (ADM) system predicts a repayment probability of $0.3$ and hence rejects Amina's application.
Amina appeals, saying, ``If the ADM's predictive model were based on data from a world without gender discrimination, my application would have been successful because my repayment probability would be $0.97$ instead of $0.3$. My estimated PS is $-0.67$ with a $95\%$ confidence interval (CI) of $(-0.9, -0.55)$. This suggests I was significantly discriminated against due to my gender.''
The first row of Figure \ref{fig:iml_viz} shows the PS and its CI. 
The blue and orange bars below 
show
the PS contributions (with CIs) of the global and individual intercepts and two features: 
\textit{STEM} (binary, degree in STEM vs.\ other) and \textit{Lead} (binary, job with leadership role vs.\ other). This would allow the bank to argue against unlawful gender discrimination by arguing that the PS is driven by an effect mediated by career path -- which, in their view, is a valid reason for higher credit risk.
However, the significant negative individual intercept suggests a gender effect that goes beyond these admissible paths, and thus 
would substantiate Amina's claim of unlawful
gender discrimination
(see Section \ref{sec:iml} for details). 



\paragraph{Auditing ADM} Chimamanda works for a government regulatory agency and is tasked with auditing the bank's ADM system. She examines the subgroup PS of various groups and has two main findings: (i) in the group of people who wear glasses, the distribution of PS is not centered around zero. 
Since wearing glasses is not a PA, this finding is reported to the bank for further monitoring without immediate consequences; 
(ii) in the group of Black women, the distribution of PS is also not centered around zero, but is significantly shifted. Because a (fictitious) law states that for protected groups, the PS must be centered around 0, 
the ADM system is not approved for operational use. 

\paragraph{Auditing real-world DGP} Chimamanda's colleague Kimberlé is tasked with auditing the NYPD's policing strategy regarding its ``Stop, Question and Frisk'' program. It is suspected that this strategy may suffer from bias at the intersection of race and gender. Kimberlé uses an analysis similar to Chimamanda's, with the difference that she does not test a given ADM system, i.e., a given ML model, but the actual data-generating process (DGP) that led to the data at hand, i.e., the (black-box) policing strategy of NYPD. To do this, she first trains an ML model on the real data and audits this as a surrogate for the true DGP. 

\subsection{Related Work}
\label{sec:related-work}

The proposed PS are intended for addressing a (presumed to be) non-neutral status quo.
We position our work relative to other fairML works that aim at changing such status quo by implementing, according to 
\citet{wachter_bias_2021}, ``bias-transforming'' methods \citep[e.g.,][]{kusner_counterfactual_2017, black_fliptest_2020, alvarez_counterfactual_2023}.
\citet{kusner_counterfactual_2017} famously introduce counterfactual fairness, which compares the factual distribution to its counterfactual counterpart \citep[using the steps of abduction, action, and prediction of][]{PearlCausality2009} in which the downstream influence of the PA is accounted for all other attributes.
\citet{black_fliptest_2020} and \citet{alvarez_counterfactual_2023}, respectively, test for individual discrimination by comparing the observed profiles to generated ones in which the seemingly neutral attributes are updated conditional on the effect of the PA.
In terms of fairML methods, our work relates mainly to pre-processing methods such as those proposed by \citet{plecko_fair_2020} and \citet{bothmann_causal_2023}.
We differ from all these works by explicitly formulating what drives the non-neutrality of the status quo in the form of privilege.

Noteworthy are the ongoing fairML discussions on achieving long-term fairness \citep[e.g.,][]{DBLP:conf/www/HuC18, DBLP:conf/fat/DAmourSABSH20, DBLP:conf/fat/SchwobelR22} and addressing the accuracy-fairness trade-off  \citep[e.g.,][]{DBLP:conf/nips/WickpT19, DBLP:journals/corr/abs-2011-03173, DBLP:journals/natmi/RodolfaLG21, leininger-tradeoffs-2025}, respectively.
The former argues that fairness interventions need to consider their impact over time. 
The latter argues that the trade-off is trivial as long as the data used for training is biased. 
Together with \citet{wachter_bias_2021} and its bias-preserving versus bias-transforming distinction, 
both discussions are examples of a more explicit direction within fairML of using algorithmic tools to intervene in the status quo.
Our work adds privilege to the discussion, viewing it as a consequence of the non-neutral status quo.

\section{Background and Notation}
\label{sec:background}


\paragraph{FiND World}

For constructing a ``fair world'', we follow the philosophical rationale of \citet{bothmann_what_2024} who propose a fictitious, normatively desired (FiND) world, where the PAs have no direct nor indirect causal effect on the target. 
We present the classification case in the remainder, but  an extension to regression is straightforward.
The basis for a decision or treatment, also called  ``task-specific merit'', is the individual probability\footnote{Individual in the sense that we condition on all of a person's characteristics, measured and unmeasured -- which we suppress notationally for convenience.} $\pi = \P(Y=1)$ for target $Y \in \{0,1\}$ in the real world, if no PAs are present.\footnote{E.g., probability of paying back the mortgage.} 
Under presence of PAs, the task-specific merit is taken to be the counterpart $\psi = \P(Y_F=1)$ in the FiND world with $Y_F \in \{0,1\}$ denoting the target there, i.e., while the label spaces are identical, distributions can differ.
Using the task-specific merit as basis for decisions, ``equals are treated equally and unequals are treated unequally'', where equality is measured either in the real or the FiND world, depending on normative stipulations regarding PAs. 
\citet{leininger-tradeoffs-2025} show that several common fairness notions -- on the group and on the individual level -- are simultaneously fulfilled in the FiND world, overcoming the fairness-accuracy trade-off and the impossibility theorem \cite{chouldechova_fair_2017,kleinberg_inherent_2017}.\footnote{\citet{bothmann_what_2024} and \citet{leininger-tradeoffs-2025} elaborate on differences between this approach and counterfactual fairness.}

\paragraph{Approximating the FiND World}
\label{sec:methods}

In practice, we do not have access to this FiND world and  approximate it empirically. This means that all descendants $\xv$ of the PAs have to be mapped or ``warped'' to values $\xtil$ that approximate their counterfactual values $\xv_F$ in the FiND world (on training data, this includes warping the target $y$ to $\yt$). 
The corresponding task-specific merit in this ``warped world'' is denoted by $\pitil = \P(\tilde{Y} = 1)$ with warped-world target $\tilde{Y} \in \{0,1\}$. Probabilities $\pi, \pitil, \psi$ in the different worlds are approximated by functions $\pix, \pitx, \psix: \Xspace \rightarrow [0,1]$ of observable features $\xv, \xtil, \xv_F \in \Xspace$. These, in turn, are estimated by ML models $\pixh, \pitxh, \psixh: \Xspace \rightarrow [0,1]$.

Such ``warping'' methods may or may not be based on the concept of causality. 
Warping methods rooted in causality are better-suited to approximate the FiND world, yet come with the obvious drawback of requiring a directed acyclic graph (DAG). 
When the DAG is misspecified, the success of the methods is questionable (and will be investigated empirically in Section \ref{sec:experiments}). 
In the experiments below, we focus on two methods, namely fairadapt by \citet{plecko_fair_2020} and a residual-based warping (later referred to as ``res-based warping'') proposed by \citet{bothmann_causal_2023}. 
Both are causal methods and very closely adopt the philosophy of the FiND world concept 
\citep[see also][]{bothmann_causal_2023,leininger-tradeoffs-2025}. 
A thorough comparison with other warping methods is left for future research. 

Since PS -- as presented below -- are a general class of scores, the number of possible estimation methods is not limited. 
The general concept of PS is agnostic to the precise definition of the fair world. It only requires the ability to (approximately) predict in both this world and the real one. 



\section{Privilege Scores (PS)}
\label{sec:formalization}

The PS $\priv$ quantifies the task-specific privilege of an individual. 
%
To derive $\priv$ we need to compare real-world treatment with fair-world treatment. 
Since the treatment is not based on the true values $\pi$ and $\psi$ but on the values $\pix$ and $\psix$ computable by the features, we define a PS as the privilege of an individual resulting from computing the decision basis in the real world instead of computing it in the FiND world:\footnote{\citet{bothmann_what_2024} introduce a treatment function $\sfu$ transforming task-specific merit normatively to define a treatment. For ease of presentation, we assume this to be the identity.} 

\begin{definition}[Privilege Score]
A privilege score $\priv: \Xspace \times \Xspace \rightarrow [-1,1]$ is a comparison of the treatment of an individual in the real versus in a fair world. It is a function of the individual's feature vector $\xv$ 
in the real world and $\xv_F$ in the fair world. 
More explicitly, it is the difference
\[\priv = \priv(\xv, \xv_F) = \pix - \psix\]
of the respective probabilities.\footnote{We could also use the ratio but argue that the difference enhances interpretability later.}
\end{definition}


We estimate $\priv$ by estimating $\pif$ and $\psif$ (1) and approximating the FiND world via warping (2):
\begin{align*}
    \priv &= \pix - \psix \\ 
    &\stackrel{(1)}{\approx} \pixh - \psixh \\ 
    &\stackrel{(2)}{\approx} \pixh - \pitxth =\hp.
\end{align*}

\paragraph{Theoretical analysis}

Analyzing the bias of the estimator $\hp$, we can derive:
\begin{align*}
    \E(\hp) &\stackrel{\text{if models are unbiased}}{=} \pix - \pitil(\xtil) & \\
    &\stackrel{\text{if warped = FiND world}}{=} \pix - \psix & = \priv. 
\end{align*}

With unbiased ML models $\pih$ and $\pitilh$, and with a perfect warping method, $\hp$ is an unbiased estimator for $\priv$. Analyzing the variance of the estimator $\hp$, we can derive:
\begin{align*}
    \var(\hp)&= \var(\pixh) + \var(\pitxth) \\
    & \quad - 2 \cov(\pixh, \pitxth).
\end{align*}
This means, to minimize the variance, we need low-variance ML models $\pih$ and $\pitilh$ which should be correlated, i.e., real and warped world should be close. However, we do not want to push the warped world toward the real world, as we want to have a good approximation of the FiND world for unbiasedness. 
In summary, we want (i) unbiased ML models $\pih$ and $\pitilh$, (ii) low-variance ML models $\pih$ and $\pitilh$, (iii) very good approximation of FiND world via warped world, (iv) and variance is smaller if the gap between real and warped world is smaller (which means -- if we are not willing to sacrifice unbiasedness -- variance is smaller if real and FiND world are closer, i.e., if the real world is ``fairer''). 

\paragraph{Uncertainty Quantification}
\label{sec:uq}

We can use bootstrapping (or subsampling) to construct CIs for PS. 
Algorithm \ref{alg:bootstrap-any} describes a general bootstrapping algorithm for a combined prediction pipeline. 
In the experiments considered later, the learning step consists of (i) learning the warping on bootstrapped data $\D_b$, (ii) warping $\D_b$ to $\Dt_b$, (iii) learning the models $\pih_b$ and $\pitilh_b$ based on $\D_b$ and $\Dt_b$, respectively, and the prediction step consists of (i) warping the test observation $\xv$ to $\xtil$, (ii) computing the predictions $\pih_b(\xv)$ and $\pitilh_b(\xtil)$.
However, the general algorithm also allows, e.g., to include the step of discovering the DAG first during learning, making it even more general.
The resulting CI is:
\[\cih_B = \left[\hp_{(\alpha/2)}, \hp_{1-(\alpha/2)}\right],\]
where $\hp_{(\alpha/2)}$ and $\hp_{1-(\alpha/2)}$ are the 
respective quantiles of the scores $\hp_b$, 
and $B$ is the number of bootstrap iterations. It can be interpreted as $\P(\cihi_B \ni \privi) = 1 - \alpha$. 

\begin{algorithm}
\caption{Bootstrapping PS}
\label{alg:bootstrap-any}
\begin{algorithmic}[1]
\STATE {\bfseries Input:} Train data $\D$, 
test observation $\xi$
\FOR{$b = 1, \dots, B$}
    \STATE Draw bootstrap sample $\D_b$ from $\D$
    \STATE Learn prediction pipeline on $\D_b$
    \STATE Apply prediction pipeline on $\xi$ 
    \STATE Compute score $\hpi_b = \pih_b(\xi) - \pitilh_b(\xti)$
\ENDFOR
\STATE \textbf{Output:} $\hpi_1, \dots, \hpi_B$
\end{algorithmic}
\end{algorithm}

\section{Privilege Score Contributions (PSCs)}
\label{sec:iml}

We aim at three types of interpretability tasks: (i) local interpretability to explain why a particular individual PS is low/high, 
(ii) global interpretability to explain the global impact of individual features on PS, 
(iii) find and describe subgroups that have unusually low/high PS. 
If we interpret the PS $\hpx$ as a black-box ML model, we can use the entire toolbox of model-agnostic interpretable ML (IML). 
We can ``ignore'' the complexity of having two ML models and the warping between the two worlds, 
i.e., we can interpret PS $\hpx$ in terms of real world features $\xv$, see also \ref{app:iml-standard}.
%
%

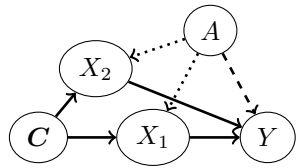
\begin{figure}[ht]
    \centering

        \begin{tikzpicture}[x=3in, y=4in] 
        \node[ellipse, draw] (v0) at (0.6,-0.24) {$Y$};
        \node[ellipse, draw] (v1) at (0.2,-0.24) {$\bm{C}$};
        \node[ellipse, draw] (v2) at (0.4,-0.24) {$X_1$};
        \node[ellipse, draw] (v4) at (0.3,-0.15) {$X_2$};
        \node[ellipse, draw] (v5) at (0.5,-0.10) {$A$};

        \draw [line width=1pt, ->] (v1) edge (v2);
        \draw [line width=1pt, ->] (v1) edge (v4);
        \draw [line width=1pt, ->] (v2) edge (v0);
        \draw [line width=1pt, ->] (v4) edge (v0);

        \draw [line width=1pt, dashed, ->] (v5) edge (v0); 
        \draw [line width=1pt, dotted, ->] (v5) edge (v2);
        \draw [line width=1pt, dotted, ->] (v5) edge (v4);
        \end{tikzpicture}
        
    \caption{Exemplary DAGs. Real world: all arrows exist. Warped world: only solid arrows exist. PSCs quantify the effect of breaking the dotted and dashed arrows separately.}
    \label{fig:DAG-example-psc}
\end{figure}

However, in many use cases, we would like to add transparency by being able to explicitly interpret the effect of warping on PS.
We cannot use an off-the-shelf IML method to do this, so we propose a novel method, called \textbf{Privilege Score Contributions (PSCs)}. 
What appears to be one privilege -- e.g., gender privilege -- may actually consist of several privileges -- e.g., gender may influence not only career choice (through societal norms) but also income for a given job (through direct discrimination). 
For local interpretations of PS, we want to separate the effect of removing a particular privilege by removing the corresponding causal effect starting in the PA, from the effect of using warped world instead of real world data for training.
To this end, we propose PSCs, that quantify the contributions of each single privilege to the total PS. 
This is equivalent to fictitiously ``breaking the wheel'' of discrimination by ``breaking'' the causal effect of each PA-related mediator path separately.
Let us take fictitious data for Amina who is applying for a mortgage and let us assume, the bank's ADM system predicts the probability of repayment. Figure \ref{fig:DAG-example-psc} shows exemplary DAGs for the real and warped world, and Table \ref{tab:shap_example} shows Amina's values for both worlds.\footnote{Note that these DAGs only serve illustrational purposes and true DAGs will likely be more complex.} 
We want to split the PS of $\hpx = -0.67$ into different contributions.

\begin{table}[h]
\centering
\caption{Fictitious data. $A$ is the binary PA \textit{Gender} (male (1) or female (0)); $C$ is a confounder \textit{Age}; $X_1$ and $X_2$ are binary features and reflect Amina's choices regarding study program (STEM field (1) or not (0)) and regarding job type (leadership role (1) or not (0)), respectively; $\tilde{X}_1$ and $\tilde{X}_2$ are warped versions; $\pixh$ and $\pitxth$ describe predictions of repayment $Y$ (yes (1) or no (0)) in the real and warped world; $\hpx$ is the estimated PS.}
\begin{tabular}{c|c|c|c|c|c|c|c|c}
$A$ & $C$ & $X_1$ & $\tilde{X}_1$ & $X_2$ & $\tilde{X}_2$ & $\pixh$ & $\pitxth$ & $\hpx$ \\
\hline
0   & 22   & 0   & 0.6   & 0   & 0.2   & 0.30   & 0.97   & -0.67 \\
\end{tabular}
\label{tab:shap_example}
\end{table}

\paragraph{Definition of PSCs}

(Estimated) PS can be written as
\begin{equation}
    \label{eq:psc-1}
    \hpx = \pixh - \pitxth = \underbrace{\pixh - \pih(\xtil)}_{\sum_{j=1}^k\gx[j]} + \underbrace{\pih(\xtil) - \pitxth}_{\pxtil[0]},
\end{equation}
where $k$ is the number of privileges induced by the PA or the number of PA-related mediator paths, which is equal to the number of arrows starting in the PA and leading to features (i.e., the dotted arrows in Figure \ref{fig:DAG-example-psc}).\footnote{The 
dashed arrow $A\rightarrow Y$ is treated separately: We do not warp $Y$ at prediction time as we do not know it. However, the intercept term can be interpreted as the part of the PS that can not be attributed to the other $k$ privileges, and hence reflects the contribution of using warped-world instead of real-world training data, i.e., the effect that goes beyond warping the individual feature set which is attributable to the differing worlds.} 
This decomposition splits the PS into (i) the local \textbf{privilege score contribution} $\gx[j]$, i.e., the effect that arrow/privilege $j$ has for the real-world model 
by warping its descending features, and (ii) the effect of using real-world instead of warped-world training data on the warped features as well as interaction effects of privileges with warped features -- the \textbf{local intercept} $\pxtil[0] = \pih(\xtil) - \pitxth$. Alternatively, we can go to the warped model first and analyze the effect that warping the features has on the warped-world model:
\begin{equation}
    \label{eq:psc-2}
    \hpx = \pixh - \pitxth = \underbrace{\pitilh(\xv) - \pitxth}_{\sum_{j=1}^k\tgx[j]} + \underbrace{\pixh - \pitilh(\xv)}_{\px[0]},
\end{equation}
While both alternatives can be computed equally well and may be interesting in a practical use case, we prefer the first option: 
After warping, the feature space in the warped world is likely to have smaller ranges in some dimensions than the original feature space in the real world, because the PA effects have been eliminated and thus the possible heterogeneity due to the PA is reduced. 
This could lead to situations where the feature vectors are outside the distribution of the training data used for the warped world model, leading to artificially noisy predictions if a learner is chosen that does not extrapolate well.

The local intercept $\pxtil[0]$ describes the local difference between real and warped world models for the warped feature vector. 
We introduce a general level shift by splitting $\pxtil[0]$:
\begin{align*}
    \hpx &= \underbrace{\left(\pih(\xtil) - \pitxth\right)}_{\pxtil[0]} + \sum_{j=1}^k\gx[j] \\
    &= \underbrace{(\pih(\xtil) - \bar{\pih}) + (\bar{\pih} - \bar{\pitilh}) + (\bar{\pitilh} - \pitilh(\xtil))}_{\pxtil[0]} 
    + \sum_{j=1}^k\gx[j] \\
    &= \underbrace{(\bar{\pih} - \bar{\pitilh})}_{\priv_g}
    + \underbrace{(\pih(\xtil) - \bar{\pih}) - (\pitilh(\xtil) - \bar{\pitilh})}_{\priv_{\xtil}}  
    + \sum_{j=1}^k\gx[j], \\
\end{align*}
where averages $\bar{\pih}$ and $\bar{\pitilh}$ are computed for training data, using the real world feature values.
The \textbf{global intercept} $\priv_g$ describes a general, global shift between the real world and the warped world that applies to the entire population.
The \textbf{individual intercept} $\priv_{\xtil}$ 
centers the local intercept around the global intercept, which facilitates its interpretation.

\paragraph{Framing via Shapley values}

We can frame our approach in terms of Shapley values \cite{shapley_value_1953} using the specific value function from Eq.\ \ref{eq:value}, motivated by the goal of fairly distributing (possibly interacting) effects on PS across PA-related mediator paths.
A coalition $S \subseteq P$ is a set of arrows starting in the PA and ending in features, $P=\{1, \dots, k\}$ is the set of all these $k$ arrows (dotted arrows in Figure \ref{fig:DAG-example-psc}). 
%
The \textbf{value functions} $v(S)$ in the real world is defined as:
\begin{equation}
\label{eq:value}
v(S) = \pixh - \pih(\xv_{S}),    
\end{equation}
where $\xv_{S}$ is the feature vector after having removed all and only the arrows in $S$, i.e., features on respective paths have been warped. This is a value function since
\[v(\emptyset) = \pixh - \pixh = 0,\]
and
\[v(P) = \pixh - \pih(\xtil)\] 

\textbf{Marginal contribution of privilege $j \in P$} is defined as:
\[ v(S \cup \{j\}) - v(S) = \pih(\xv_S) - \pih(\xv_{S \cup \{j\}}).\]
This leads to the \textbf{Shapley-style PSCs}:
\begin{equation}
\label{def:psc}
\gx[j] = \frac{1}{M}\sum_{m=1}^M\pih(\xv_{S_j^{(m)}}) - \pih(\xv_{S_j^{(m)} \cup \{j\}}),
\end{equation}
where $S_j^{(m)}$ is the set of players that appear before $j$ in permutation $m$, and $M=k!$ is the total number of permutations of $P$.
%
We show efficiency (Theorem \ref{th:eff}) and other axioms of fair payouts in \ref{app:iml-efficiency}.
\begin{theorem}[Efficiency]
\label{th:eff}
Let $\{\gx[j]\}_{j \in P}$ be the Shapley values induced by $v$. Then
$$
\sum_{j=1}^k \gx[j]
\;=\;
v(P)
\;=\;
\pih(\xv) - \pih(\xtil).
$$
\end{theorem}
Analogously, we can define value function $\tilde{v}(S)$ and Shapley-Style PSCs $\tgx[j]$ in the warped world using $\tilde{v}(S) = \pitilh(\xv) - \pitilh(\xv_{S})$, etc.
For estimation, we do not have the computational complexity problem of Shapley values, since we usually consider a rather small number $k$ (task-specific justifiable PA-related mediator paths). 
Similar to Algorithm \ref{alg:bootstrap-any}, we can compute \textbf{bootstrap CIs} for any PSC by performing the PSC computation on bootstrap samples. 

\paragraph{Partially warped features}

In a DAG such as in Figure \ref{fig:DAG-example-psc}, the vector $\xv_{S}=(\xtil_{d(S)}, \xv_{nd(S)})$ consists of warped versions of features $d(S)$ on paths descending from $S$ and real versions of non-descending features $nd(S)$. However, there can be DAGs with features descending from more than one privilege. Appendix \ref{app:iml-partial-warping} treats these cases.

\paragraph{Downstream analyses}

PSCs can be used for other use cases, adding to the above individual interpretation:
\begin{itemize}
    \item Global strength of privileges: We can analyze (moments of) distributions of PSCs 
    to investigate the global effect of the respective privilege. By this means, we can quantify a privilege, which vice versa quantifies by how much discrimination can be reduced via tackling the respective privilege with political actions. E.g.: If $A \rightarrow X_1$ induces an average privilege of increasing/decreasing the risk score by $\zeta$, tackling the gender-related discrepancy of career choice can diminish the gender-related discrepancy in risk scores by this $\zeta$. Furthermore, the intercept terms hint at direct discrimination, as they capture the part of the PS which is not directly attributable to single privileges.
    \item Feature importance: By averaging the absolute PSCs for each privilege, we can derive task-specific ``privilege importances''. E.g., it might be the case that the gender pay gap is a major driving factor for gender-related discrepancies in credit risk scores, but not for gender-related discrepancies in recidivism scores.
    \item Investigate interactions between privilege and features: SHAP dependence plots \cite{lundberg_local_2020} can be used to visualize the effect that a feature $x_l$ has on $\gx[j]$. This could be used to examine whether gender-based privilege via income depends on age or race, revealing subgroups of the PA that suffer more from privilege (such as young Women of Color) than others.
    \item Subgroup analysis: We can investigate subgroups that on average show exceptionally high or low PS values, e.g., via bump hunting \cite{friedman_bump_1999}.
\end{itemize}

\paragraph{Standard Shapley values}
Alternatively, we can use ``standard'' Shapley values for interpretation (see Appendix \ref{app:iml-shapley}). Interpretation of the Shapley values $\eta_j$ differs substantially from the interpretation of PSCs while both angles may be interesting in a practical use case. Shapley values $\eta_j$ focus on relating the PS of an individual to their features and disregard the question of whether that contribution is due to the warping of the feature or due to the differing model in the warped world. On the other hand, PSCs focus on the PA effects and disentangle the effect of privilege $j$ ($\gx[j]$) from the effect of using the warped-world model instead of the real-world model ($\pxtil[0]$). 

\paragraph{Examples of PSCs}
Figure \ref{fig:iml_viz} visualizes PSCs for the example of Amina, corresponding to Table \ref{tab:shap_example}: 
The PSC of $\hgx[1]=-0.25$ from \textit{STEM} can be interpreted as follows: Warping Amina's value of \textit{STEM} from $0$ to $0.6$ (i.e., fictitiously increasing her probability of having chosen a STEM career) increases the real-world prediction by $0.25$. 
The PSC of $\hgx[2]=-0.12$ means: Warping Amina's value of \textit{Lead} from $0$ to $0.2$ (i.e., fictitiously increasing her probability of having a job with leadership role) increases her real-world prediction by $0.12$.
Using the warped-world prediction model instead of the real-world prediction model increases the prediction further by $0.3$ (the local intercept of $\hpxtil[0] = -0.3$ can be splitted into an individual intercept of $\hp_{\xv}=-0.4$ 
and a global intercept of $\hp_g=0.1$). 
 
To put it another way: 
All people get -- on average -- a higher prediction in the warped world ($\hp_g$), but those with features in a local neighborhood of Amina's warped values get a lower  prediction ($\hpxtil[0]$). 
The warping of her feature values has the effect 
of increasing her prediction in the real world.
Bootstrap CIs indicate that while the intercepts and the PSC of \textit{STEM} are significantly non-zero, the PSC of \textit{Lead} is not significantly deviating from $0$. 
In this fictitious example, we would see (i) a gender-related negative privilege mediated by career choice (\textit{STEM}) -- because of the significant $\hgx[1]$, and (ii) a further gender-related negative privilege not attributable to \textit{STEM} or \textit{Lead},  
i.e., hinting at other mediators or direct discrimination -- because of the significant $\hpxtil$.


\section{Experiments}
\label{sec:experiments}

In this section, we analyze the proposed framework of PS empirically. We investigate the behavior of PS estimators for different settings of a simulation study 
and examine the practical applicability 
on real-world datasets.\footnote{Code is available as supplementary material.}

\subsection{Simulated data}
\label{sec:exp-sim}

The goal of the simulation study is to compare the candidate methods (fairadapt and res-based warping, see Section \ref{sec:background}) for estimating PS in order to judge their suitability for the real-world experiments.
%
%
We pursue the following research questions: (RQ1) How biased are the methods? What MSE do the methods have? What coverage do the CIs have? 
(RQ2) How sensitive are the methods to misspecification of the DAG?
(RQ3) Can PSC derive explanations that mirror correctly the true privileges?



We investigate two different scenarios: (SC) In a fictional mortgage lending example we use the DAG in Figure \ref{fig:DAG-example-psc} again; 
(SM) Misspecification: Data are generated with an additional effect $A \rightarrow C$ while DAG as in (SC) is used for estimation. For each scenario, we draw $n=1000$ observations in $M=50$ iterations, see \ref{app:sec:sim-setup} for full setup.

\paragraph{Results}
We highlight some results here and show full results in Appendix \ref{app:sec:sim-results}. 
%
%
%
For (RQ1), Table \ref{tab:sim_study_s2s3} (SC, left) shows that estimates of PS with fairadapt and res-based warping are approximately unbiased and have low MSE. The coverage of the CIs for both methods is slightly above the confidence level of $90\%$, meaning that while they can be used for significance statements, there is room for more power. For (RQ2), Table \ref{tab:sim_study_s2s3} (SM, right) shows that misspecification degrades the performance, while res-based warping seems more robust than fairadapt.

\begin{table}[!h] 
    \centering
        \caption{PS: Mean (5\% and 95\% quantiles) of bias, MSE and $(1-\alpha)$-CI coverage, $\alpha=0.1$.} 
    \resizebox{0.49\textwidth}{!}{ 
        \begin{tabular}[t]{lcccc}
            \toprule
            \multicolumn{1}{c}{ } & \multicolumn{2}{c}{SC} & \multicolumn{2}{c}{SM} \\
            \cmidrule(l{3pt}r{3pt}){2-3} \cmidrule(l{3pt}r{3pt}){4-5}
            Metric & fairadapt & res-based & fairadapt & res-based\\
            \midrule
            Bias & -0.001 & 0.002 & 0.037 & 0.001\\
             & (-0.015, 0.012) & (-0.014, 0.014) & (0.013, 0.062) & (-0.012, 0.016)\\
            MSE & 0.011 & 0.012 & 0.024 & 0.017\\
             & (0.007, 0.015) & (0.007, 0.016) & (0.014, 0.036) & (0.012, 0.023)\\
            Coverage & 0.977 & 0.969 & 0.94 & 0.949\\
             & (0.944, 0.994) & (0.939, 0.988) & (0.897, 0.976) & (0.93, 0.963)\\
            \bottomrule
        \end{tabular}
    }
    \label{tab:sim_study_s2s3}
\end{table}

For (RQ3), we compare true PSCs with estimated PSCs for (SC), see Table \ref{tab:rq3} (and \ref{tab:app-sim-sm-psc}). 
Again, estimates are approximately unbiased and have low MSE. Coverage of the CIs are within sensible ranges, only coverage of PSC $\gx[2]$ for \textit{Saving} using res-based warping is slightly too low. 


\begin{table}[!h] 
    \centering
        \caption{PSC -- scenario (SC): Metrics as in Table \ref{tab:sim_study_s2s3}. Top: res-based, bottom: fairadapt.} 
    \resizebox{0.49\textwidth}{!}{ 
        \begin{tabular}[t]{lcccc}
            \toprule
            Metric & $\priv_g$ & $\priv_{\xtil}$ & $\gx[1]$ & $\gx[2]$\\
            \hline
            Bias & 0.002 & -0.002 & -0.001 & 0.003 \\ 
            & (-0.007, 0.012) & (-0.013, 0.009) & (-0.009, 0.009) & (-0.001, 0.006) \\ 
            MSE & 0 & 0.011 & 0.002 & 0.002 \\ 
            & (0, 0) & (0.007, 0.016) & (0.001, 0.004) & (0.001, 0.002) \\ 
            Coverage & 0.973 & 0.984 & 0.958 & 0.866 \\ 
            & (0.817, 1) & (0.973, 0.996) & (0.936, 0.981) & (0.846, 0.884) \\ 
            \hline
            Bias & -0.001 & 0 & -0.003 & 0.002 \\ 
            & (-0.008, 0.007) & (-0.008, 0.008) & (-0.013, 0.008) & (-0.001, 0.005) \\ 
            MSE & 0 & 0.01 & 0.002 & 0.001 \\ 
            & (0, 0) & (0.007, 0.014) & (0.001, 0.004) & (0.001, 0.002) \\ 
            Coverage & 1 & 0.983 & 0.953 & 0.923 \\ 
            & (1, 1) & (0.97, 0.996) & (0.928, 0.973) & (0.89, 0.952) \\ 
            \hline
        \end{tabular}
    }
    \label{tab:rq3}
\end{table}

We conclude that both methods can be used to estimate PS/PSC in real-world applications, while future work could investigate how they perform on more complex DAGs.
As res-based warping seems to be more robust against misspecification, we report real-world results for it in the next section and refer to Appendix \ref{app:sec:exp-real} for fairadapt results.

\subsection{Real-world data}
\label{sec:exp-real}

We analyze data from the Home Mortgage Disclosure Act (HMDA) and from the Law School data.\footnote{See Appendix \ref{app:sec:exp-real} for details on the setup and more results.} Data are splitted into $80\%$/$20\%$ train/test sets and results reported on test set.

\paragraph{Mortgage Data}

\begin{figure}[ht]
    \centering
    \begin{subfigure}{0.2\textwidth}
        \centering
        \begin{tikzpicture}[x=3in, y=4in] 
        \node[ellipse, draw] (v0) at (0.6,-0.24) {$Y$};
        \node[ellipse, draw] (v1) at (0.2,-0.24) {$\bm{C}$};
        \node[ellipse, draw] (v2) at (0.5,-0.31) {$X_1$};
        \node[ellipse, draw] (v3) at (0.4,-0.24) {$X_2$};
        \node[ellipse, draw] (v4) at (0.3,-0.17) {$X_3$};
        \node[ellipse, draw] (v5) at (0.52,-0.12) {$A$};

        \draw [line width=1pt, ->] (v1) edge (v2);
        \draw [line width=1pt, ->] (v1) edge (v3);
        \draw [line width=1pt, ->] (v1) edge (v4);
        \draw [line width=1pt, ->] (v2) edge (v0);
        \draw [line width=1pt, ->] (v3) edge (v0);
        \draw [line width=1pt, ->] (v4) edge (v0);

        \draw [dashed, line width=1pt, ->] (v5) edge (v0); 
        \draw [dotted, line width=1pt, ->] (v5) edge (v2);
        \draw [dotted, line width=1pt, ->] (v5) edge (v3);
        \draw [dotted, line width=1pt, ->] (v5) edge (v4);
        \end{tikzpicture}
        \caption{}
        \label{fig:DAG-mortgage}
    \end{subfigure}
    \hfill
    \begin{subfigure}{0.2\textwidth}
        \centering
        \begin{tikzpicture}[x=3in, y=4in] 
        \node[ellipse, draw] (v0) at (0.4,-0.6) {UGPA}; 
        \node[ellipse, draw] (v2) at (0.2,-0.7) {$A$};
        \node[ellipse, draw] (v4) at (0.4,-0.7) {$Y$};
        \node[ellipse, draw] (v6) at (0.4,-0.8) {LSAT};

        \draw [line width=1pt, ->] (v0) edge (v4);
        \draw [dotted, line width=1pt, ->] (v2) edge (v0); 
        \draw [dotted, line width=1pt, ->] (v2) edge (v6); 
        \draw [line width=1pt, ->] (v6) edge (v4); 
        \draw [dashed, line width=1pt, ->] (v2) edge (v4);
        \end{tikzpicture}
        \caption{} 
        \label{fig:DAG_lawschool}
    \end{subfigure}
    \caption{DAGs for (a) mortgage and (b) law school data.}
\label{fig:DAG-real-world-exp}
\end{figure}
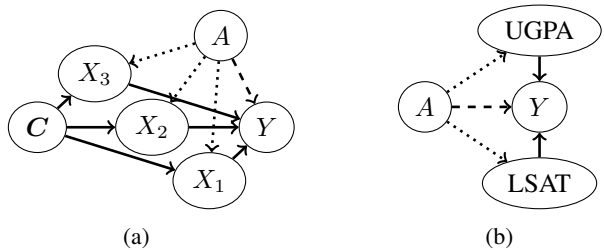

Figure \ref{fig:DAG-mortgage} shows assumed DAGs. Data consist of: binary PA \textit{Race} ($A$, white/non-white), numerical feature \textit{Amount} ($X_1$), binary features \textit{Debt} ($X_2$, debt-to-income ratio smaller than $36\%$ or not) and \textit{Purpose} ($X_3$, home purchase/other),  confounders $\bm{C}$, consisting of binary \textit{Sex} (male/female) and binary \textit{Age} (above $62$ or not) -- and binary target \textit{Action} ($Y$, loan originated or not). 
We apply res-based warping (with logit models for binary features and a gamma model for \textit{Amount}) and fairadapt for three locations: We show results with res-based warping for Louisiana (LA) here and refer to Appendix \ref{app:sec:exp-real} for results on New York county (NY) and Wisconsin (WI) and for fairadapt.





\begin{figure}[ht]
    \centering
    \includegraphics[width=0.5\textwidth]{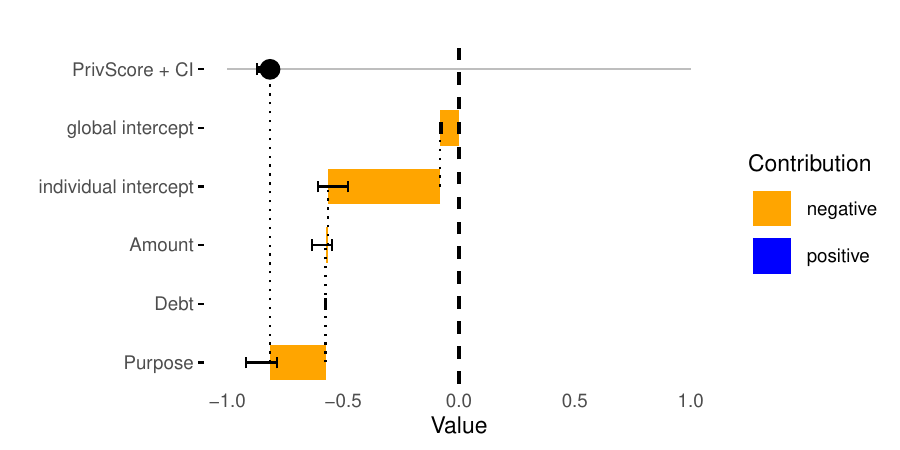} 
    \caption{LA: PS and PSCs for a non-white female person, (ID 244, see Appendix Table \ref{tab:app_la} for real and warped values).}
    \label{fig:psc_mortgage}
\end{figure}

Data from year 2022 has $n=14.758$. 
Figure \ref{fig:psc_mortgage} shows PS and PSCs for an individual (ID 244) with  high negative PS; locally, race-related negative privileges via \textit{Purpose} as well as the individual intercept seem significantly non-zero.
Globally, a regression of all PS (in the test set) on all features reveals a highly significant 
race effect of around $0.22$, meaning that ceteris paribus, the expected PS of a white applicant is additively $0.22$ higher than that of a non-white applicant, indicating a strong racial bias for LA.
For NY and WI, the race effects are also significant but  lower at $0.04$ and $0.1$, respectively.

Table \ref{tab:mortgage} summarizes PS and PSCs for the non-white subgroup ($A=0$): We see a considerable imbalance of PS $\hp$ toward negative values and the most important path of privilege seems to be the individual intercept $\priv_{\xv}$. This indicates that non-observed features could be the main mediators of discrimination or that there could be a direct race discrimination effect. Comparing features, \textit{Purpose} and \textit{Amount} have the highest importance and the bias regarding \textit{Debt} seems negligible. Note that future work has yet to develop formal tests or CI's to allow for significance statements
.

\begin{table}[ht]
\centering
\caption{Louisiana: Mean and quantiles ($\alpha=0.05$) of PS/PSC and PSC importance on test data.}
\begin{tabular}{lrrr}
  \hline
 & Mean & Quantiles $(\alpha, 1-\alpha)$ & Importance \\ 
  \hline
$\hp$ & -0.249 & (-0.815, 0.037) & -- \\ 
   $\priv_g$ & -0.082 & (-0.082, -0.082) & 0.082 \\ 
  $\priv_{\xv}$ & -0.101 & (-0.634, 0.326) & 0.269 \\ 
  $\gx[1]$ & -0.033 & (-0.135, 0.019) & 0.039 \\ 
  $\gx[2]$ & -0.010 & (-0.098, 0.000) & 0.010 \\ 
  $\gx[3]$ & -0.024 & (-0.247, 0.183) & 0.060 \\ 
   \hline
\end{tabular}
\label{tab:mortgage}
\end{table}

\paragraph{Law school data}
Figure \ref{fig:DAG_lawschool} shows assumed DAGs. We use binary PA \textit{Race} ($A$, Black/non-Black), and binary outcome $Y$ (bar exam passed or not), for $n=22.407$ observations, see \ref{app:sec:exp-real:law} for detailed setup.
Table \ref{tab:psc-law-fairadapt-sex} shows that, most interestingly, the path via LSAT ($\gx[2]$) contributes the most to Black students' negative privilege regarding the probability of passing the bar -- while intercept effects (indicators of direct effects or missing mediators) are comparatively small. This means that the negative privilege of black students can be explained mainly by differences in the LSAT, which in turn implies that policies aimed at effectively increasing racial equality should aim at equalizing LSAT scores. Further research could focus on why LSAT is more important than UGPA. 
See also Appendix \ref{app:sec:exp-real:law} for additional results.

\begin{table}[ht] 
    \centering
    \caption{Law School: Mean and quantiles ($\alpha=0.05$) of PS/PSC and PSC importance on test data} 
    \begin{tabular}{lccc}
        \hline
        Feature & Mean & Quantiles $(\alpha, 1-\alpha)$ & Importance \\ 
        \hline
$\hp$ & -0.149 & (-0.403, 0.006) & -- \\ 
   $\priv_g$ & -0.005 & (-0.005, -0.005) & 0.005 \\ 
  $\priv_{\xv}$ & -0.010 & (-0.086, 0.018) & 0.026 \\ 
  $\gx[1]$ & -0.022 & (-0.052, 0.002) & 0.023 \\ 
  $\gx[2]$ & -0.111 & (-0.285, -0.001) & 0.111 \\ 
   \hline
    \end{tabular}
    \label{tab:psc-law-fairadapt-sex}
\end{table}


\paragraph{Discussion}
The real-world data experiments demonstrated the applicability of our method. 
For the mortgage analysis, 
significant negative privilege was found
at the individual level; 
its main drivers appear to be factors not included in the data set, as the individual intercept has the highest value. 
We can also quantify and explain privilege
at the global level, 
concluding: (i) for mortgage, racial discrimination manifests itself through unobserved features. Comparing different US locations, racial bias is twice as high in LA as in WI and even five times as high as in NY; 
(ii) for law school, the racial privilege can mainly be explained by a path via LSAT, a finding which in turn can inform policy makers aiming at mitigating racial disparities.

While these experiments show the relevance of our method, we also see  limitations. In a practical use case, we would like to make significance statements about PSC importance. We leave it to future work to develop appropriate tests. Our method assumes a given DAG, and simulations have shown that misspecification of the DAG affects the robustness of the estimates. In an applied study, great care needs to be taken in defining this DAG, ideally in collaboration with domain experts. A worthwhile direction for future work would also be to investigate how causal discovery methods can be added to our framework.

\section{Conclusion and Outlook}
\label{sec:discussion}


We proposed privilege scores and presented a general framework for their estimation, including uncertainty quantification. In addition, we presented privilege score contributions that can be used to explain how privileges -- related to given protected attributes -- are constituted in real-world applications. Experiments on simulated and real-world data compared the estimation methods and demonstrated their practical applicability. 

We see several opportunities for future work: other (causal and non-causal) estimation methods that could populate our general framework can be investigated. Rigorous tests of PSC importance would allow for significance claims. A user study would investigate how feasible PS and PSCs are for practitioners. Finally, PS could be used in model training to produce fairer models.

\section*{Impact Statement}


This paper presents work whose goal is to advance the field of 
machine learning. There are many potential societal consequences 
of our work, none which we feel must be specifically highlighted here.

\bibliography{mybib-zotero-modified, ReferencesFromJose, mybib-manually}
\bibliographystyle{xxxxxx2025}

\newpage
\appendix
\onecolumn

\section{Details on interpretability methods}

\subsection{Example for applying standard IML method on PS}
\label{app:iml-standard}
E.g., for feature effects we can use ICE or PD plots, for feature importance we can use different variants of permutation feature importance, SAGE values, or LOCO \citep[see e.g.,][for an overview]{ewald_guide_2024}. 
As just one example, Algorithm \ref{alg:pfi} describes how to compute permutation feature importances for PS. 

\begin{algorithm}[ht]
\caption{Permutation Feature Importance (PFI)}
\label{alg:pfi}
\begin{algorithmic}[1]
\STATE Learn warping and ML models in real and warped world ($\pih()$ and $\pitilh()$)

\STATE Compute test error of predictions $\hp$
\FOR{each feature $x_j$}
    \STATE Permute the values of feature $x_j$
    \STATE Compute predictions $\hp_*$ using same warping and models $\pih()$ and $\pitilh()$
    \STATE Compute test error of predictions $\hp_*$
    \STATE Estimate FI via difference of the test errors
\ENDFOR
\IF{multiple permutations are performed}
    \STATE Repeat the permutation process several times and average
\ENDIF
\end{algorithmic}
\end{algorithm}

\subsection{PSC proofs}
\label{app:iml-efficiency}

\subsubsection{Setup and Definition of PSC}

Let $P = \{1,\dots,k\}$ be the set of $k$ privilege-inducing arrows (players). 
For a subset $S \subseteq P$, define
$$
v(S) \;=\; \pih(\xv) \;-\; \pih(\xv_S),
$$
where $\xv$ is the original real-world feature vector, and $\xv_S$ is the same feature vector with \emph{exactly} the arrows in $S$ ``removed'' (warped).  By construction:
$$
v(\varnothing) \;=\; \pih(\xv) \;-\; \pih(\xv) \;=\; 0
\quad\text{and}\quad
v(P) \;=\; \pih(\xv) \;-\; \pih(\xtil)
$$

We aim to allocate $v(P)$ among the $k$ players (arrows) in a principled way.

Two equivalent definitions of Shapley values exist \cite{strumbelj_explaining_2014}:

\medskip

\textbf{Set-based Shapley Values for Player $j \in P$:}
$$
\gx[j] \;=\;
\sum_{S \subseteq P \setminus \{j\}}
\frac{|S|!\,\bigl(k - |S| -1\bigr)!}{k!}
\;
\bigl[v(S \cup \{j\}) - v(S)\bigr].
$$

\medskip

\textbf{Order-based Shapley Values for Player $j \in P$:}

A \emph{permutation} of $P$ is any ordering $\pi = (\pi(1), \pi(2), \dots, \pi(k))$.  
Let $\mathrm{S}_\pi(j)$ be the set of players that appear \emph{before} $j$ in the permutation $\pi$, i.e., $\mathrm{S}_\pi(j) = \{\pi(1),\dots,\pi(r-1)\}$ if $\pi(r)=j$. 
Then the \textbf{Shapley value} of player $j$ is
$$
\gx[j] 
\;=\;
\frac{1}{k!}\;
\sum_{\pi \in \mathfrak{S}_P}
\Bigl[v\!\bigl(\mathrm{S}_\pi(j) \cup \{\,j\}\bigr)
\;-\;
v\!\bigl(\mathrm{S}_\pi(j)\bigr)\Bigr],
$$
where $\mathfrak{S}_P$ is the set of all $k!$ permutations of $P$.  
In words, we look at the “marginal contribution” of $j$ each time it ``arrives'' in a permutation (given whichever players arrived before it), and then average over all permutations.

\subsubsection{Axioms of fair payouts}


While \textbf{efficiency} is explicitly proved below, the other properties directly follow from the definition of PSCs as Shapley values:

\begin{itemize}
    \item \textbf{Symmetry}: If two arrows $i, j \in P$ make the same marginal contributions in every coalition $S \subseteq P\setminus\{i,j\}$, they trivially receive identical PSC values.
    \item \textbf{Dummy Property}: An arrow that marginally contributes nothing to any coalition, i.e. $v(S \cup \{j\}) \;=\; v(S)$ for all $S \subseteq P\setminus\{j\}$, obtains a PSC of $0$.
\end{itemize}

\subsubsection{Efficiency Proof: Set-based Shapley Values}


\begin{proofsketch}

\noindent 

\textbf{Step 1: Split the sum into pieces.}
$$
\sum_{j=1}^k \gx[j] 
\;=\;
\sum_{j=1}^k 
\;\sum_{\,S \subseteq P\setminus\{j\}}
\;\frac{|S|!\,\bigl(k - |S| -1\bigr)!}{k!}
\;\Bigl[\,v\bigl(S \cup \{j\}\bigr) - v(S)\Bigr].
$$
Each pair $(j,S)$ contributes a difference $v(S \cup \{j\}) - v(S)$. Let $T = S \cup \{j\}$.

Additional Remark: Notice that the coefficient
  $$
    \frac{|S|!\,(k - |S| -1)!}{k!}
  $$
  can be viewed as the probability (over all $k!$ permutations) that $S$ is exactly the set of predecessors of $j$. 
  This viewpoint will help in understanding later cancellation.

\medskip
\noindent
\textbf{Step 2: Identify how $v(P)$ is counted once overall.}

\begin{itemize}
    \item Whenever $T = P$, we must have $j \in P$ and $S = P \setminus \{j\}$. There are exactly $k$ such pairs $(j,S)$ because $j$ can be any of the $k$ elements in $P$.
    \item For each such pair, $\lvert S\rvert = k-1$, the coefficient becomes
    $$
       \frac{(k-1)!\,(k - (k-1) - 1)!}{k!}
       \;=\;
       \frac{(k-1)!\,(0)!}{k!}
       \;=\;
       \frac{1}{k}.
    $$
    \item Hence each of the $k$ occurrences of $v(P)$ is multiplied by $\tfrac{1}{k}$, summing to $\bigl(\frac{1}{k}\times k\bigr)=1$. Thus \emph{overall}, $v(P)$ is counted once in total. 
\end{itemize}

\medskip
\noindent
\textbf{Step 3: Show that all $v(T)$ with $T \neq P$ and $T \neq \emptyset$ cancel.}

 Each nonempty $T \subset P$ appears \emph{positively} in $v(T) - v(T \setminus \{j\})$ (because of $+\,v(T)$), 
  and also \emph{negatively} in the larger set $v\bigl(T \cup \{j'\}\bigr) - v(T)$ (as $-\,v(T)$). 
  Crucially, the factorial coefficients match up so that the total weight of all positive appearances of $v(T)$ 
  equals the total weight of all negative appearances of $v(T)$. 
  Hence they completely cancel out.
  
More precisely, for each nonempty $T \subset P$:
\begin{itemize}
    \item \textbf{Positive occurrence.} 
    If $T = S \cup \{j\}$ for some $j \in T$, then $v(T)$ appears \emph{positively} in the difference 
    $
      v(S \cup \{j\}) \;-\; v(S).
    $

    \item \textbf{Negative occurrence.} 
    If $T \subset P$ is not the full set, then there exists at least one element $j' \notin T$. We can thus form the larger set $T \cup \{j'\}$. In the Shapley sum, this larger set contributes
    $
      v\bigl(T \cup \{j'\}\bigr) \;-\; v(T),
    $
    which contains $-\,v(T)$ as the \emph{negative} appearance of $v(T)$.

    \item \textbf{Multinomial coefficient.} 
    Fix a nonempty $T$. Let $|T|$ denote its size. Inside the sum, $T$ appears once for each $j \in T$ with $S = T \setminus \{j\}$. The coefficient in front of $v(T)$ for such a pair $(j,S)$ is
    $$
      \frac{|T\setminus\{j\}|!\,\bigl(k - |T\setminus\{j\}| -1\bigr)!}{k!}
      \;=\;
      \frac{(|T|-1)!\,(k - |T|)!}{k!}.
    $$
    Since there are $|T|$ possible ways to choose $j\in T$, multiplying by $|T|$ yields
    $$
      |T| 
      \;\cdot\; 
      \frac{(|T|-1)!\,(k - |T|)!}{k!}
      \;=\;
      \frac{|T|!\,(k - |T|)!}{k!}
      \;=\;
      \frac{k!}{k!}
      \;=\;
      1.
    $$
    Where the identity $|T|!\,(k - |T|)! = k!$ holds because to permute $k$ distinct elements, 
we can imagine first choosing which $|T|$ are in one group and ordering them 
(in $|T|!$ ways), then ordering the remaining $k - |T|$ (in $(k - |T|)!$ ways). 
Overall, this accounts for all $k!$ permutations. 
    Therefore, each nonempty $T$ receives a total \emph{positive} weight of exactly $+1$ when summing over all $j \in T$.

    \item \textbf{Cancellation.} 
    Every nonempty $T \neq P$ also appears \emph{negatively} as part of a larger set’s difference, ensuring one positive and one negative occurrence of $v(T)$. Consequently, these contributions cancel each other out, so $v(T)$ does not remain in the final sum unless $T=P$.
\end{itemize}
\medskip
\noindent
\textbf{Step 4: Conclusion.}

Since $v(\varnothing) = 0$ does not contribute and $v(P)$ remains once in total, we get
$$
\sum_{j=1}^k \gx[j] 
\;=\;
\underbrace{v(P)}_{\text{one net positive}} 
\;+\; 
\underbrace{\sum_{T \neq P} \bigl[v(T)\ \text{terms cancel}\bigr]}_{0} 
\;=\;
v(P).
$$
\qedhere
\end{proofsketch}

\subsubsection{Efficiency Proof: Order-Based Shapley Values}

\begin{proofsketch}
\noindent

\textbf{Step 1: Sum the Shapley values over all players.}

$$
\sum_{j=1}^k \gx[j] 
\;=\;
\sum_{j=1}^k
\frac{1}{k!}
\sum_{\pi \in \mathfrak{S}_P}
\Bigl[v\bigl(\mathrm{S}_\pi(j)\cup\{j\}\bigr)
- 
v\bigl(\mathrm{S}_\pi(j)\bigr)\Bigr]
=
\frac{1}{k!}
\sum_{\pi \in \mathfrak{S}_P}
\sum_{j=1}^k
\Bigl[v\bigl(\mathrm{S}_\pi(j)\cup\{j\}\bigr)
- 
v\bigl(\mathrm{S}_\pi(j)\bigr)\Bigr].
$$

\medskip
\noindent
\textbf{Step 2: Telescoping within each permutation.}

Fix a particular permutation $\pi$.  List its elements in order:
$$
\bigl(\pi(1), \pi(2), \dots, \pi(k)\bigr).
$$
Within this permutation, the inner sum
$\sum_{j=1}^k
\bigl[v(\mathrm{S}_\pi(j)\cup\{j\}) - v(\mathrm{S}_\pi(j))\bigr]$
can be viewed as a chain of marginal contributions:
$$
v\bigl(\{\pi(1)\}\bigr) - v(\varnothing)
\;+\;
v\bigl(\{\pi(1), \pi(2)\}\bigr) - v\bigl(\{\pi(1)\}\bigr)
\;+\;\dots\;+\;
v\bigl(\{\pi(1),\ldots,\pi(k)\}\bigr) - v\bigl(\{\pi(1),\ldots,\pi(k-1)\}\bigr).
$$
All intermediate terms telescope, leaving exactly
$$
v\bigl(\{\pi(1),\ldots,\pi(k)\}\bigr)
\;-\;
v(\varnothing)
\;=\;
v(P)
\;-\;
0
\;=\;
v(P).
$$

\medskip
\noindent
\textbf{Step 3: Average over all permutations.}

Since every permutation $\pi$ yields exactly $v(P)$ in the telescoped sum, we have
$$
\sum_{j=1}^k \gx[j]
\;=\;
\frac{1}{k!}
\sum_{\pi \in \mathfrak{S}_P} 
v(P)
\;=\;
\frac{1}{k!} \times \bigl(k! \cdot v(P)\bigr)
\;=\;
v(P).
$$
Hence
$$
\sum_{j=1}^k \gx[j] 
\;=\;
v(P).
$$
\end{proofsketch}





\subsection{Partially warped features}
\label{app:iml-partial-warping}

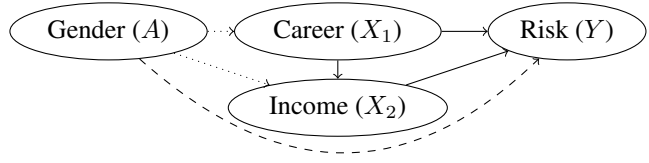
\begin{figure}[ht]
    \centering
\begin{tikzpicture}[x=3in, y=4in] 
\node[ellipse, draw] (v0) at (0.5,-0.6) {Career ($X_1$)}; 
\node[ellipse, draw] (v2) at (0.1,-0.6) {Gender ($A$)};
\node[ellipse, draw] (v4) at (0.9,-0.6) {Risk ($Y$)};
\node[ellipse, draw] (v6) at (0.5,-0.7) {Income ($X_2$)}; 

\draw [->] (v0) edge (v4);
\draw [->] (v0) edge (v6);
\draw [dotted, ->] (v2) edge (v0); 
\draw [dotted, ->] (v2) edge (v6); 
\draw [->] (v6) edge (v4); 
\draw [dashed, ->] (v2) .. controls (0.4,-0.8) and (0.625,-0.8) .. (v4); 

\end{tikzpicture}
    \caption{Exemplary DAG for partial warping.}
    \label{fig:DAG-example-psc-partial}
\end{figure}

In a DAG such as in Figure \ref{fig:DAG-example-psc}, the vector $\xv_{S}=(\xtil_{d(S)}, \xv_{nd(S)})$ consists of warped versions of features $d(S)$ on paths descending from $S$ and real versions of non-descending features $nd(S)$. However, there can be DAGs like the one shown in Figure \ref{fig:DAG-example-psc-partial}, with features descending from more than one privilege. In the example, Income is affected by both the privilege that directly affects Income and the privilege that indirectly affects Income through career choice. For such cases, we introduce the notion of ``partially warping'' a feature: To derive the PSC of Income, we have to consider the $M=2$ cases of (i) having or (ii) not having the arrow $A \rightarrow X_1$. In the first case, we consider the effect of privilege $A \rightarrow X_2$ with $A \rightarrow X_1$ still present (i.e., $x_1$ is not warped), and for Eq. \ref{def:psc} compute
\[\pih(\xv_{S_j^{(1)}}) - \pih(\xv_{S_j^{(1)} \cup \{j\}}) = \pih(x_1, x_2) - \pih(x_1, \xt_2(x_2)), \]
in the second case, we look at the effect of privilege $A \rightarrow X_2$ without $A \rightarrow X_1$ (i.e., $x_1$ is warped to $\xt_1$)
\[\pih(\xv_{S_j^{(2)}}) - \pih(\xv_{S_j^{(2)} \cup \{j\}}) = \pih(\xt_1, \xt_2(x_1)) - \pih(\xt_1, \xt_2), \]
where $\xt_2(x_1)$ means that $x_2$ was partially warped through the path via $x_1$, and $\xt_2 = \xt_2(x_1, x_2)$.

This does not change the definition of PSCs in Eq.\ \ref{def:psc}, as the number of summands remains the same, and this is just a finer-grained definition of $\xv_{S}$. Since the current warping methods are not yet fully capable of performing these partial warpings out-of-the-box, in the experiments below we limit ourselves to cases that do not require partial warping, emphasizing that this is not a shortcoming of PSCs, but future work in the field of warping method development.

\subsection{Shapley values}
\label{app:iml-shapley}
Alternatively, we can use ``standard'' Shapley values for interpretation. Shapley values $\beta_j$ for the real world can be defined as

\[\pixh = \underbrace{\E_\xv(\pixh)}_{\beta_0} + \sum_{j=1}^p \beta_j,\]

where shapley values $\tilde{\beta}_j$ in the warped world can be defined as

\[\pitxth = \underbrace{\E_{\xtil}(\pitxth)}_{\tilde{\beta}_0} + \sum_{j=1}^p \tilde{\beta}_j.\]

Bringing both together, shapley values $\eta_j$ of the PS are defined via

\begin{align*}
    \hpx &= \pixh - \pitxth\\
    &=\underbrace{(\beta_0-\tilde{\beta}_0)}_{\eta_0} + \sum_{j=1}^p\underbrace{(\beta_j-\tilde{\beta}_j)}_{\eta_j}.
\end{align*}

Due to additivity, this defines consistent shapley values for the PS, which also means that standard estimation techniques can be used.

\section{Simulation study}
\label{app:sec:exp-sim}

\subsection{Simulation study -- setup}
\label{app:sec:sim-setup}

For our simulation study, we generate $M=50$ synthetic data sets each of size $n=1000$, using the DAG shown in Figure \ref{fig:DAG-example-psc}.
Data consist of: binary PA \textit{Gender} ($A$, female/male), numerical feature \textit{Amount} ($X_1$), binary feature \textit{Saving} ($X_2$, low/high), numerical confounder \textit{Age} ($C$) -- and binary target \textit{Risk} ($Y$, (1) for low and (0) high). 
To create the bootstrap confidence intervals for PS and PSC, we draw $B=100$ samples. For each of the two experimental scenarios (SC) and (SM), we sample from a separate SCM both of which are provided below. For both real and FiND world we draw random samples using the same structural assignments with the only difference that in the FiND world, the value of the PA is forced to take the value of the advantaged group which is the male group, denoted with (1). To ensure we draw the same individual for each world, we use the same random seed for both worlds and reset it before starting simulation from each world. 

Structural assignments from the (SC) simulation scenario are defined as follows:
\begin{equation*}
    \begin{aligned}
    A &\sim \text{Binomial}(p = 0.69), \\
    C &\sim \text{Gamma}(k = 9.76, \theta = 3.64), \\
    X_1|A,C &\sim \text{Gamma}\left(k = 0.74^{-1}, \theta = 0.74 \cdot \exp\left(7.9 + 0.175 \cdot A + 0.005 \cdot C\right)\right), \\
    X_2|A,C &\sim \text{Binomial}\left(\text{probit}\left(4 - 1.25 \cdot A - 0.1 \cdot C\right)\right), \\
    Y|A,C,X_1,X_2 &\sim \text{Binomial}\left(\text{probit}\left(0.9 + 0.1 \cdot C + 1.75 \cdot A - 0.7 \cdot X_2 - 0.001 \cdot X_1 \right)\right),
\end{aligned}
\end{equation*}

where $k$ denotes the shape parameter and $\theta$ the scale parameter of the gamma distribution. Equivalently, the SCM used to generate data for the (SM) simulation scenario consist of the parameters
\begin{equation*}
    \begin{aligned}
    A &\sim \text{Binomial}(p = 0.69), \\
    C|A &\sim \text{Gamma}\left(k = 10, \theta = 2 \cdot \exp\left(0.1 + 0.8 \cdot A\right)\right), \\
    X_1|A,C &\sim \text{Gamma}\left(k = 0.74^{-1}, \theta = 0.74 \cdot \exp\left(7.9 + 0.175 \cdot A + 0.005 \cdot C\right)\right), \\
    X_2|A,C &\sim \text{Binomial}\left(\text{probit}\left(4 - 1.25 \cdot A - 0.1 \cdot C\right)\right), \\
    Y|A,C,X_1,X_2 &\sim \text{Binomial}\left(\text{probit}\left(0.9 + 0.1 \cdot C + 1.75 \cdot A - 0.7 \cdot X_2 - 0.001 \cdot X_1\right)\right),
    \end{aligned}
\end{equation*}

where again  $k$ and $\theta$ denote the shape and scale parameter of the gamma distribution respectively. Note that the SCM from (SM) differs from the SCM of the (SC) scenario only with respect to the structural assignment of the node $C$. However, during training and inference, we use the SCM from (SC) which thus leads to a misspecification. 

To predict outcome $Y$, we implement a random forest classifier. During training we utilize random search with 25 evaluations and 3-fold CV. For evaluation, we use 3-fold CV as outer resampling, and report metrics on test folds. We report summary statistics of the data set for one iteration of (SC) in Table \ref{tab:sim_sumstats}.

\begin{table}[!h]
    \centering
    \caption{Summary statistics for one random sample of size $n=1000$ for the simulation scenario (SC).}
    \label{tab:sim_sumstats}
    \resizebox{\ifdim\width>\linewidth\linewidth\else\width\fi}{!}{
    \begin{tabular}{lrrrrrrrrr}
        \toprule
        \multicolumn{1}{c}{ } & \multicolumn{3}{c}{All} & \multicolumn{3}{c}{Female} & \multicolumn{3}{c}{Male} \\
        \cmidrule(l{3pt}r{3pt}){2-4} \cmidrule(l{3pt}r{3pt}){5-7} \cmidrule(l{3pt}r{3pt}){8-10}
        Variable & n & mean & std. dev. & n & mean & std. dev. & n & mean & std. dev.\\
        \midrule
        Age & 1000 & 36.166 & 11.293 & 307 & 36.611 & 10.994 & 693 & 35.969 & 11.425\\
        Amount & 1000 & 3569.819 & 3049.610 & 307 & 3121.363 & 2682.451 & 693 & 3768.486 & 3180.480\\
        Risk & 1000 & 0.748 & 0.434 & 307 & 0.681 & 0.467 & 693 & 0.778 & 0.416\\
        Saving & 1000 & 0.440 & 0.497 & 307 & 0.612 & 0.488 & 693 & 0.364 & 0.481\\
        \bottomrule
    \end{tabular}}
\end{table}

\subsection{Simulation study -- additional results}
\label{app:sec:sim-results}

Scenario (SC): Table \ref{tab:app-sim-sc} shows full results for both res-based warping (top) and fairadapt (bottom).

Scenario (SM): Table \ref{tab:app-sim-sm-psc} shows full results for both res-based warping (top) and fairadapt (bottom).

\begin{table}[ht] 
    \centering
            \caption{SC -- PS/PSCs. Data are based on the simulation study with 50 iterations following (SC) simulation scenario and CIs are computed via bootstrap. First rows report the mean, second rows the 5th and 95th percentile. The upper panel presents results for res-based warping the bottom those for fairadapt.} 
            \label{tab:app-sim-sc}
        \begin{tabular}{lcccc}
            \hline
            & Bias & MSE & Coverage & Width CI \\ 
            \hline
            $\hp$ & 0.002 & 0.012 & 0.969 & 0.292 \\ 
            & (-0.014, 0.014) & (0.007, 0.016) & (0.939, 0.988) & (0.267, 0.318) \\ 
            $\priv_g$ & 0.002 & 0 & 0.973 & 0.036 \\ 
            & (-0.007, 0.012) & (0, 0) & (0.817, 1) & (0.032, 0.041) \\ 
            $\priv_{\xtil}$ & -0.002 & 0.011 & 0.984 & 0.294 \\ 
            & (-0.013, 0.009) & (0.007, 0.016) & (0.973, 0.996) & (0.269, 0.321) \\ 
            $\gx[1]$ & -0.001 & 0.002 & 0.958 & 0.09 \\ 
            & (-0.009, 0.009) & (0.001, 0.004) & (0.936, 0.981) & (0.075, 0.107) \\ 
            $\gx[2]$ & 0.003 & 0.002 & 0.866 & 0.013 \\ 
            & (-0.001, 0.006) & (0.001, 0.002) & (0.846, 0.884) & (0.007, 0.02) \\ 
            \hline
            $\hp$ & -0.001 & 0.011 & 0.977 & 0.296 \\ 
            & (-0.015, 0.012) & (0.007, 0.015) & (0.944, 0.994) & (0.265, 0.321) \\ 
            $\priv_g$ & -0.001 & 0 & 1 & 0.037 \\ 
            & (-0.008, 0.007) & (0, 0) & (1, 1) & (0.034, 0.041) \\ 
            $\priv_{\xtil}$ & 0 & 0.01 & 0.983 & 0.285 \\ 
            & (-0.008, 0.008) & (0.007, 0.014) & (0.97, 0.996) & (0.261, 0.313) \\ 
            $\gx[1]$ & -0.003 & 0.002 & 0.953 & 0.09 \\ 
            & (-0.013, 0.008) & (0.001, 0.004) & (0.928, 0.973) & (0.076, 0.109) \\ 
            $\gx[2]$ & 0.002 & 0.001 & 0.923 & 0.025 \\ 
            & (-0.001, 0.005) & (0.001, 0.002) & (0.89, 0.952) & (0.016, 0.036) \\ 
            \hline
        \end{tabular}
\end{table}

\begin{table}[ht] 
    \centering
           \caption{SM -- PS/PSCs. Data are based on the simulation study with 50 iterations following (SM) simulation scenario and CIs are computed via bootstrap. First rows report the mean, second rows the 5th and 95th percentile. The upper panel presents results for res-based warping the bottom those for fairadapt.} 
           \label{tab:app-sim-sm-psc}
        \begin{tabular}{lcccc}
          \hline
         & Bias & MSE & Coverage & Width CI \\ 
              \hline
            $\hp$ & 0.001 & 0.017 & 0.949 & 0.279 \\ 
               & (-0.012, 0.016) & (0.012, 0.023) & (0.93, 0.963) & (0.259, 0.3) \\ 
              $\priv_g$ & -0.04 & 0.002 & 0.313 & 0.059 \\ 
               & (-0.056, -0.024) & (0.001, 0.003) & (0, 0.667) & (0.045, 0.078) \\ 
              $\priv_{\xtil}$ & -0.039 & 0.032 & 0.728 & 0.3 \\ 
               & (-0.057, -0.026) & (0.024, 0.039) & (0.545, 0.862) & (0.28, 0.321) \\ 
              $\gx[1]$ & -0.018 & 0.004 & 0.924 & 0.106 \\ 
               & (-0.031, -0.004) & (0.003, 0.007) & (0.877, 0.955) & (0.092, 0.122) \\ 
              $\gx[2]$ & 0.014 & 0.002 & 0.798 & 0.011 \\ 
               & (0.011, 0.017) & (0.001, 0.002) & (0.736, 0.88) & (0, 0.03) \\ 
               \hline
            $\hp$ & 0.037 & 0.024 & 0.94 & 0.303 \\ 
               & (0.013, 0.062) & (0.014, 0.036) & (0.897, 0.976) & (0.275, 0.336) \\ 
              $\priv_g$ & -0.016 & 0.001 & 0.82 & 0.072 \\ 
               & (-0.034, 0.004) & (0, 0.001) & (0.334, 1) & (0.053, 0.091) \\ 
              $\priv_{\xtil}$ & -0.029 & 0.02 & 0.923 & 0.312 \\ 
               & (-0.04, -0.017) & (0.014, 0.03) & (0.783, 0.98) & (0.283, 0.346) \\ 
              $\gx[1]$ & -0.008 & 0.004 & 0.942 & 0.106 \\ 
               & (-0.023, 0.005) & (0.002, 0.006) & (0.912, 0.964) & (0.09, 0.12) \\ 
              $\gx[2]$ & 0.007 & 0.001 & 0.937 & 0.046 \\ 
               & (-0.001, 0.013) & (0.001, 0.002) & (0.88, 0.972) & (0.033, 0.065) \\ 
               \hline
        \end{tabular}
\end{table}

\clearpage

\section{Real-world data}
\label{app:sec:exp-real}

\subsection{Mortgage data}
\label{app:sec:mortgage}

\subsubsection{Data Setup}
For the 2022 Home Mortgage Disclosure Act (HMDA) data,\footnote{\url{https://ffiec.cfpb.gov/data-browser/}} we apply the following encoding and filtering steps\footnote{Comprehensive variable descriptions are available at: \url{https://ffiec.cfpb.gov/documentation/publications/loan-level-datasets/lar-data-fields}}:

\begin{itemize}
\item $Y$: A binary outcome variable representing whether the loan was approved (1) or not approved (0). This is derived from the original “action taken” variable, which categorizes loan statuses into eight distinct groups.
\item $A$: Binary protected attribute representing the applicant's race, with (1) White applicants and (0) other applicants.
\item $X_1$: Numerical feature for the loan amount.
\item $X_2$: Binary feature for the debt-to-income ratio, indicating whether the ratio is smaller than $36\%$ (1) or higher (0).
\item $X_3$: Binary feature indicating the loan's purpose, with (1) signifying home purchases and (0) for other purposes. The original variable consists of six categories.
\item $\bm{C}$: This variable combines two confounders -- age and sex. Age is represented as a binary variable indicating (1) age over 62 or (0) age 62 and below. Sex is also simplified into a binary form, with (1) for male and (0) for female. (Note: This binary simplification of gender is used purely for analytical simplicity and does not represent the authors' viewpoint.)
\end{itemize}

Table \ref{tab:hmda-summary} provides a summary of the three data sets used for state Lousiana (LA), state Wisconsin (WI), and New York County (NY). Already these rough summary statistics hint at LA exhibiting higher disparities in $Y$ with respect to race and gender than WI and NY.
For each locations, we split the data in $80\%$ training and $20\%$ test data, tune hyperparameters with 3-fold CV and random search with 25 evaluations on the training data, and report metrics on the test set.

\begin{table}[ht]
\centering
\caption{Summary statistics for HMDA data} 
\begin{tabular}{lrrr}
  \hline
 & LA & WI & NY \\ 
  \hline
Number of Observations &  73790 & 165004 &  14708 \\ 
  Sex Ratio (Female) & 0.4033 & 0.3776 & 0.3799 \\ 
  Race Ratio (Non-White) & 0.3287 & 0.1205 & 0.3704 \\ 
  Loan Purpose (Home Purchase) & 0.5963 & 0.3854 & 0.6322 \\ 
  Age Ratio (Above 62) & 0.1525 & 0.1731 & 0.1355 \\ 
  Debt Ratio (Low) & 0.4414 & 0.5164 & 0.5623 \\ 
  Mean Loan Amount &  193174 &  176538 & 1066191 \\ 
  Mean Action Taken (Non-White) & 0.4669 & 0.7337 & 0.7950 \\ 
  Mean Action Taken (White) & 0.7071 & 0.8367 & 0.8320 \\ 
  Mean Action Taken (Female) & 0.5618 & 0.8112 & 0.8171 \\ 
  Mean Action Taken (Male) & 0.6729 & 0.8323 & 0.8190 \\ 
  \hline
\end{tabular}
\label{tab:hmda-summary}
\end{table}

\subsubsection{Results State Louisiana}

\begin{figure}[ht]
    \centering
    \includegraphics[width=1\textwidth]{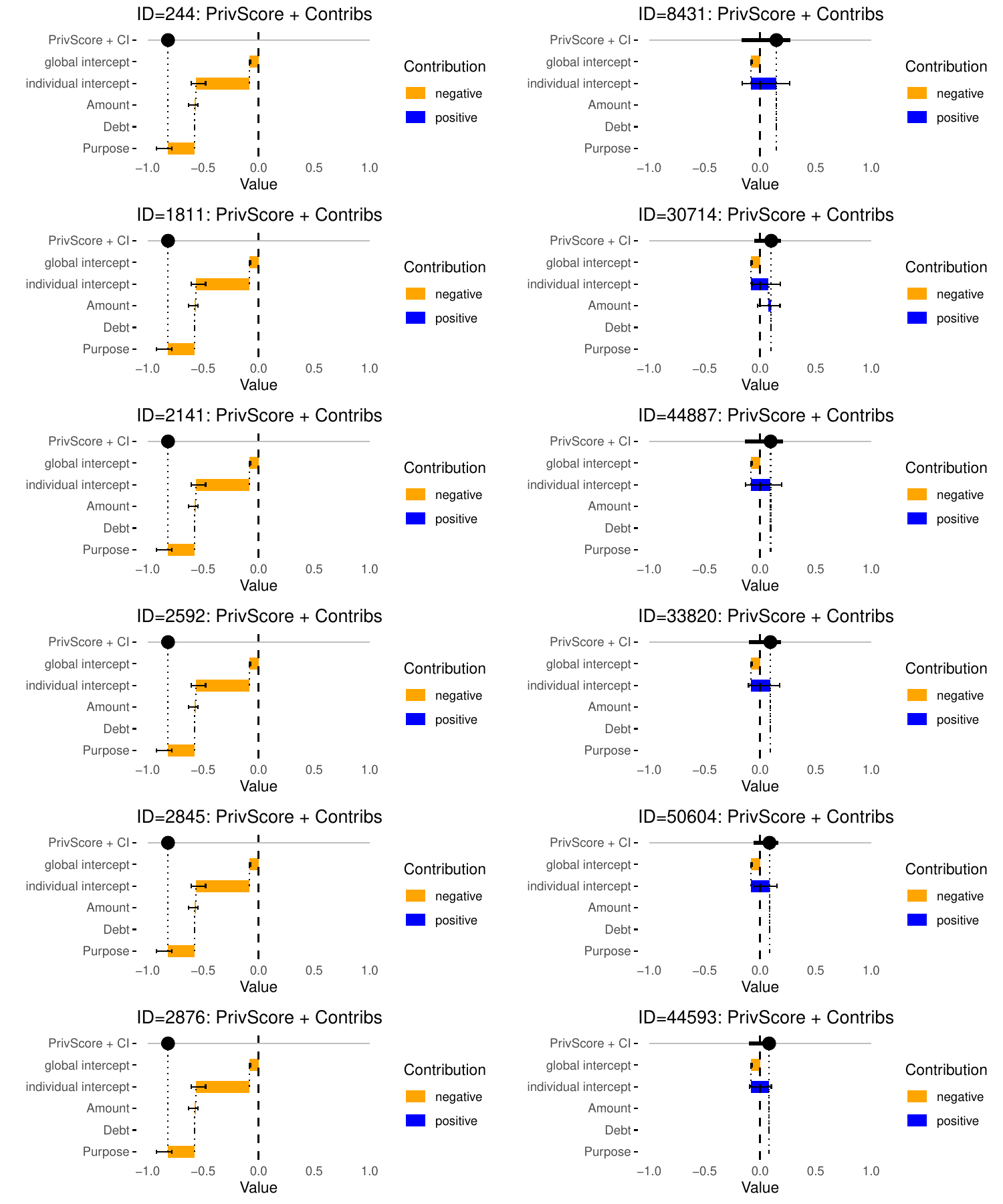} 
    \caption{LA -- res-based -- Individuals with smallest (left) and largest (right) PS.}
    \label{fig:app-la-psc-res}
\end{figure}

Table \ref{tab:app_la} shows the 6 individuals with the lowest PS and the 6 individuals with the highest PS for test data of LA using res-based warping. Figure \ref{fig:app-la-psc-res} shows PSCs for these individuals. Most notably, individuals with lowest privilege scores are young, non-white females, while there is no non-white female amongst the individuals with highest privilege. Besides the above mentioned race effect of $0.22$ on the PS, we also see a significant -- but smaller -- sex effect of $0.03$. The age effect is not significant. See Table \ref{tab:la-lm-summary} for full model summary.

\begin{figure}[ht]
    \centering
    \includegraphics[width=1\textwidth]{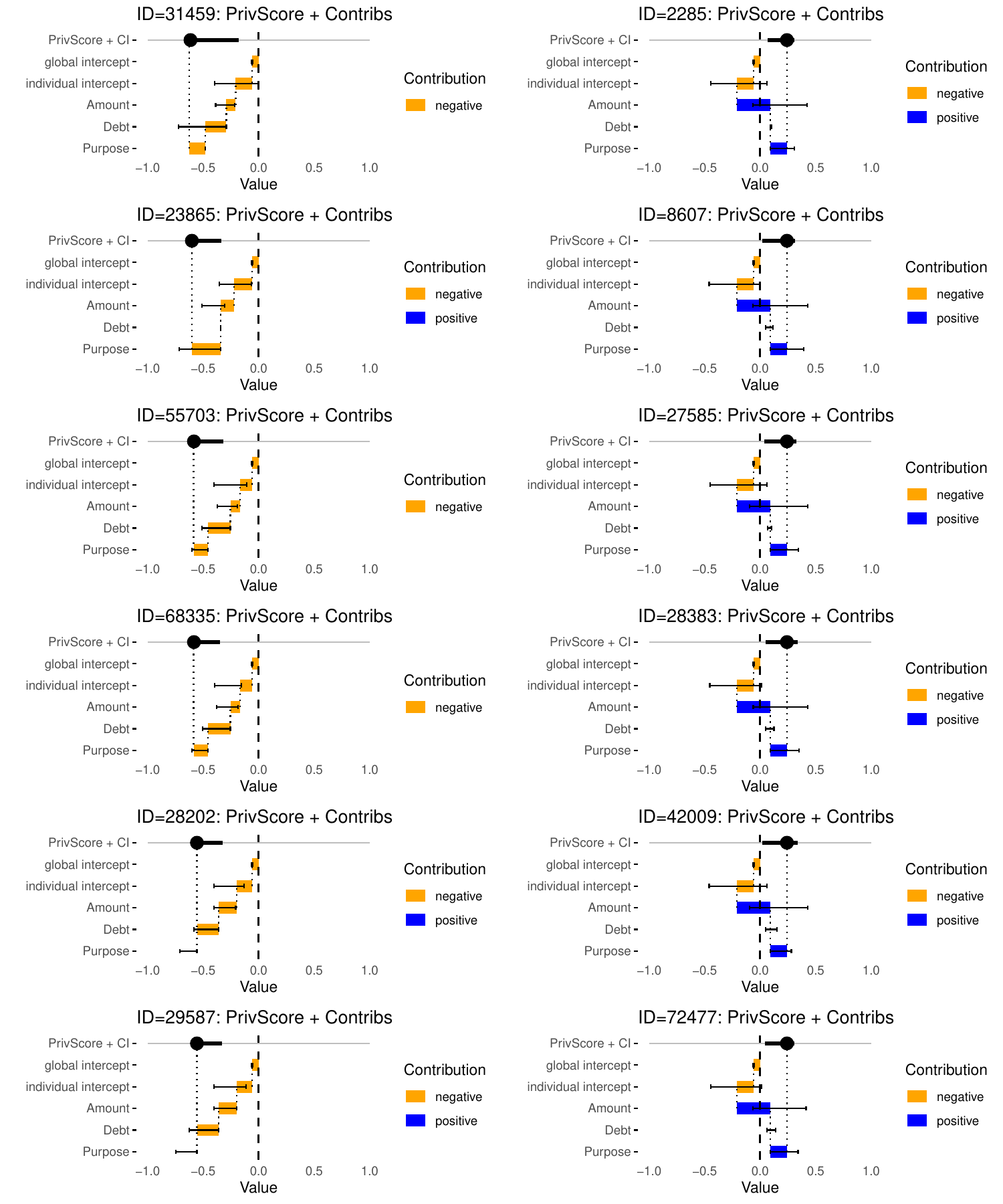} 
    \caption{LA -- fairadapt -- Individuals with smallest (left) and largest (right) PS.}
    \label{fig:app-la-psc-adapt}
\end{figure}

Table \ref{tab:app_la-adapt} shows the 6 individuals with the lowest PS and the 6 individuals with the highest PS for test data of LA using fairadapt. Figure \ref{fig:app-la-psc-adapt} shows PSCs for these individuals. 
Table \ref{tab:la-lm-summary-adapt} exhibits similar sex and race effects as for res-based warping.
Table \ref{tab:mortgage-fairadapt} shows the PS/PSC. Numbers are comparable to results of res-based warping, see Table \ref{tab:mortgage}, which substantiates the prior findings.

\begin{sidewaystable}[ht!]
\centering
\caption{LA -- res-based warping -- Individuals with smallest (upper half) and largest (lower half) PS. Warped features are indicated by ``\_w''.}
\begin{tabular}{rrrrrrrrrrrr}
  \hline
ID & sex & race & age & purpose & amount & debt & purpose\_w & amount\_w & debt\_w & pred\_real & pred\_warped \\ 
  \hline
244 &   0 &   0 &   0 &   1 & 85000 &   0 & 0.24 & 98186.99 &   0 & 0.18 & 1.00 \\ 
  1811 &   0 &   0 &   0 &   1 & 85000 &   0 & 0.24 & 98186.99 &   0 & 0.18 & 1.00 \\ 
  2141 &   0 &   0 &   0 &   1 & 85000 &   0 & 0.24 & 98186.99 &   0 & 0.18 & 1.00 \\ 
  2592 &   0 &   0 &   0 &   1 & 85000 &   0 & 0.24 & 98186.99 &   0 & 0.18 & 1.00 \\ 
  2845 &   0 &   0 &   0 &   1 & 85000 &   0 & 0.24 & 98186.99 &   0 & 0.18 & 1.00 \\ 
  2876 &   0 &   0 &   0 &   1 & 85000 &   0 & 0.24 & 98186.99 &   0 & 0.18 & 1.00 \\ 
  \hline
  8431 &   1 &   1 &   0 &   1 & 2155000 &   0 & 1.00 & 2155000.00 &   0 & 0.68 & 0.53 \\ 
  30714 &   1 &   0 &   1 &   1 & 655000 &   1 & 1.00 & 921105.10 &   1 & 0.84 & 0.74 \\ 
  44887 &   0 &   1 &   0 &   0 & 8825000 &   0 & 0.00 & 8825000.00 &   0 & 0.68 & 0.59 \\ 
  33820 &   1 &   1 &   1 &   0 & 1185000 &   0 & 0.00 & 1185000.00 &   0 & 0.64 & 0.54 \\ 
  50604 &   0 &   1 &   1 &   0 & 455000 &   0 & 0.00 & 455000.00 &   0 & 0.58 & 0.49 \\ 
  44593 &   1 &   1 &   1 &   0 & 1375000 &   0 & 0.00 & 1375000.00 &   0 & 0.66 & 0.58 \\ 
   \hline
\end{tabular}
\label{tab:app_la}
\end{sidewaystable}

\begin{sidewaystable}[ht!]
\centering
\caption{LA -- fairadapt -- Individuals with smallest (upper half) and largest (lower half) PS. Warped features are indicated by ``\_w''.}
\begin{tabular}{rrrrrrrrrrrr}
  \hline
ID & sex & race & age & purpose & amount & debt & purpose\_w & amount\_w & debt\_w & pred\_real & pred\_warped \\ 
  \hline
31459 &   0 &   0 &   1 &   1 & 105000 &   0 &   0 & 145000 &   1 & 0.24 & 0.85 \\ 
  23865 &   0 &   0 &   0 &   1 & 105000 &   1 &   0 & 145000 &   1 & 0.25 & 0.86 \\ 
  55703 &   0 &   0 &   0 &   1 & 145000 &   0 &   0 & 195000 &   1 & 0.28 & 0.86 \\ 
  68335 &   0 &   0 &   0 &   1 & 145000 &   0 &   0 & 195000 &   1 & 0.28 & 0.86 \\ 
  28202 &   0 &   0 &   0 &   1 & 145000 &   0 &   1 & 195000 &   1 & 0.28 & 0.83 \\ 
  29587 &   0 &   0 &   0 &   1 & 145000 &   0 &   1 & 195000 &   1 & 0.28 & 0.83 \\ 
    \hline
  2285 &   0 &   0 &   0 &   1 & 5000 &   0 &   0 & 25000 &   0 & 0.70 & 0.46 \\ 
  8607 &   0 &   0 &   0 &   1 & 5000 &   0 &   0 & 25000 &   0 & 0.70 & 0.46 \\ 
  27585 &   0 &   0 &   0 &   1 & 5000 &   0 &   0 & 25000 &   0 & 0.70 & 0.46 \\ 
  28383 &   0 &   0 &   0 &   1 & 5000 &   0 &   0 & 25000 &   0 & 0.70 & 0.46 \\ 
  42009 &   0 &   0 &   0 &   1 & 5000 &   0 &   0 & 25000 &   0 & 0.70 & 0.46 \\ 
  72477 &   0 &   0 &   0 &   1 & 5000 &   0 &   0 & 25000 &   0 & 0.70 & 0.46 \\ 
   \hline
\end{tabular}
\label{tab:app_la-adapt}
\end{sidewaystable}

\begin{table}[ht]
\centering
\caption{LA -- res-based -- Linear Model Summary} 
\begin{tabular}{rrrrr}
  \hline
 & Estimate & Std. Error & t value & Pr($>$$|$t$|$) \\ 
  \hline
(Intercept) & -0.3087 & 0.0041 & -75.3542 & $<$ 1e-16 \\ 
  sex & 0.0326 & 0.0032 & 10.3000 & $<$ 1e-16 \\ 
  race & 0.2200 & 0.0033 & 66.5151 & $<$ 1e-16 \\ 
  purpose & -0.0336 & 0.0032 & -10.5666 & $<$ 1e-16 \\ 
  amount & 0.0000 & 0.0000 & 22.9779 & $<$ 1e-16 \\ 
  debt & 0.0925 & 0.0031 & 30.0047 & $<$ 1e-16 \\ 
  age & 0.0003 & 0.0043 & 0.0759 & 0.93953 \\ 
   \hline
\end{tabular}
\label{tab:la-lm-summary}
\end{table}


\begin{table}[ht]
\centering
\caption{LA -- fairadapt -- Linear Model Summary} 
\begin{tabular}{rrrrr}
  \hline
 & Estimate & Std. Error & t value & Pr($>$$|$t$|$) \\ 
  \hline
(Intercept) & -0.2597 & 0.0016 & -164.0807 & $<$ 1e-16 \\ 
  sex & 0.0151 & 0.0012 & 12.3340 & $<$ 1e-16 \\ 
  race & 0.2352 & 0.0013 & 184.0681 & $<$ 1e-16 \\ 
  purpose & -0.0056 & 0.0012 & -4.5356 & 5.7903e-06 \\ 
  amount & 0.0000 & 0.0000 & 19.5736 & $<$ 1e-16 \\ 
  debt & 0.0050 & 0.0012 & 4.1613 & 3.1827e-05 \\ 
  age & 0.0063 & 0.0017 & 3.7751 & 0.00016057 \\ 
   \hline
\end{tabular}
\label{tab:la-lm-summary-adapt}
\end{table}

\begin{table}[ht]
\centering
\caption{LA -- fairadapt -- Mean and quantiles ($\alpha=0.05$) of PS/PSC and PSC importance on test data}
\begin{tabular}{lrrr}
  \hline
 & Mean & Quantiles $(\alpha, 1-\alpha)$ & Importance \\ 
  \hline
$\hp$ & -0.243 & (-0.451, -0.058) & -- \\ 
   $\priv_g$ & -0.058 & (-0.058, -0.058) & 0.058 \\ 
  $\priv_{\xv}$ & -0.119 & (-0.240, 0.005) & 0.121 \\ 
  $\gx[1]$ & -0.046 & (-0.204, 0.006) & 0.063 \\ 
  $\gx[2]$ & -0.017 & (-0.144, 0.000) & 0.017 \\ 
  $\gx[3]$ & -0.003 & (0.000, 0.000) & 0.010 \\ 
   \hline
\end{tabular}
\label{tab:mortgage-fairadapt}
\end{table}

\subsubsection{Results State Wisconsin}

\begin{figure}[ht]
    \centering
    \includegraphics[width=1\textwidth]{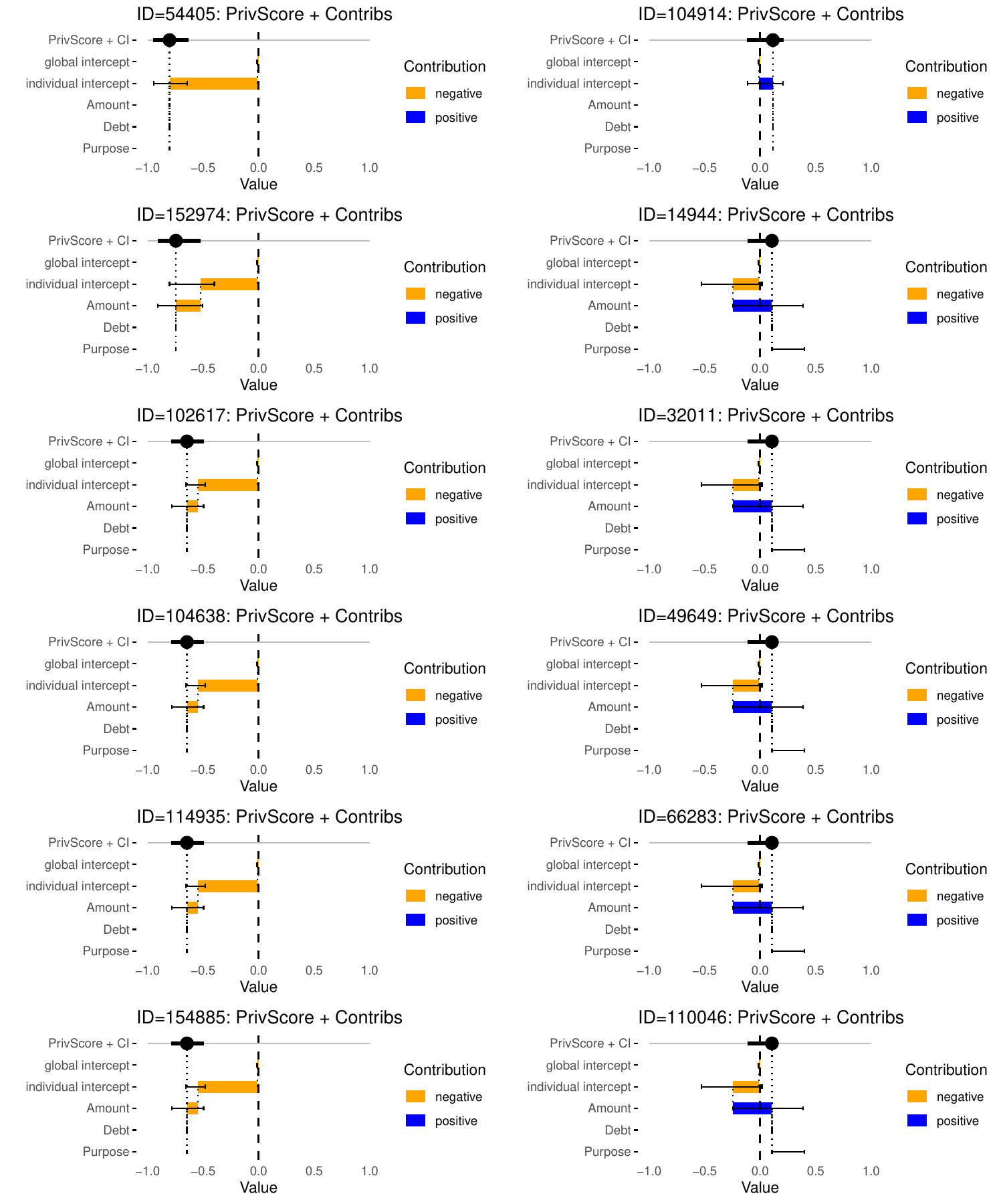} 
    \caption{WI -- res-based -- Individuals with smallest (left) and largest (right) PS.}
    \label{fig:app-wi-psc}
\end{figure}

Table \ref{tab:app_wi-res} shows the 6 individuals with the lowest PS and the 6 individuals with the highest PS for test data of WI using res-based warping. Figure \ref{fig:app-wi-psc} shows PSCs for these individuals.  
Table \ref{tab:lm_summary-wi-res} gives a model summary of regressing PS on real-world features.
Besides the above mentioned race effect of $0.10$ on the PS, we also see a significant -- but smaller -- sex effect of $0.002$. The age effect is not significant.  Table \ref{tab:psc-wi-res} shows the PS/PSC results.

\begin{figure}[ht]
    \centering
    \includegraphics[width=1\textwidth]{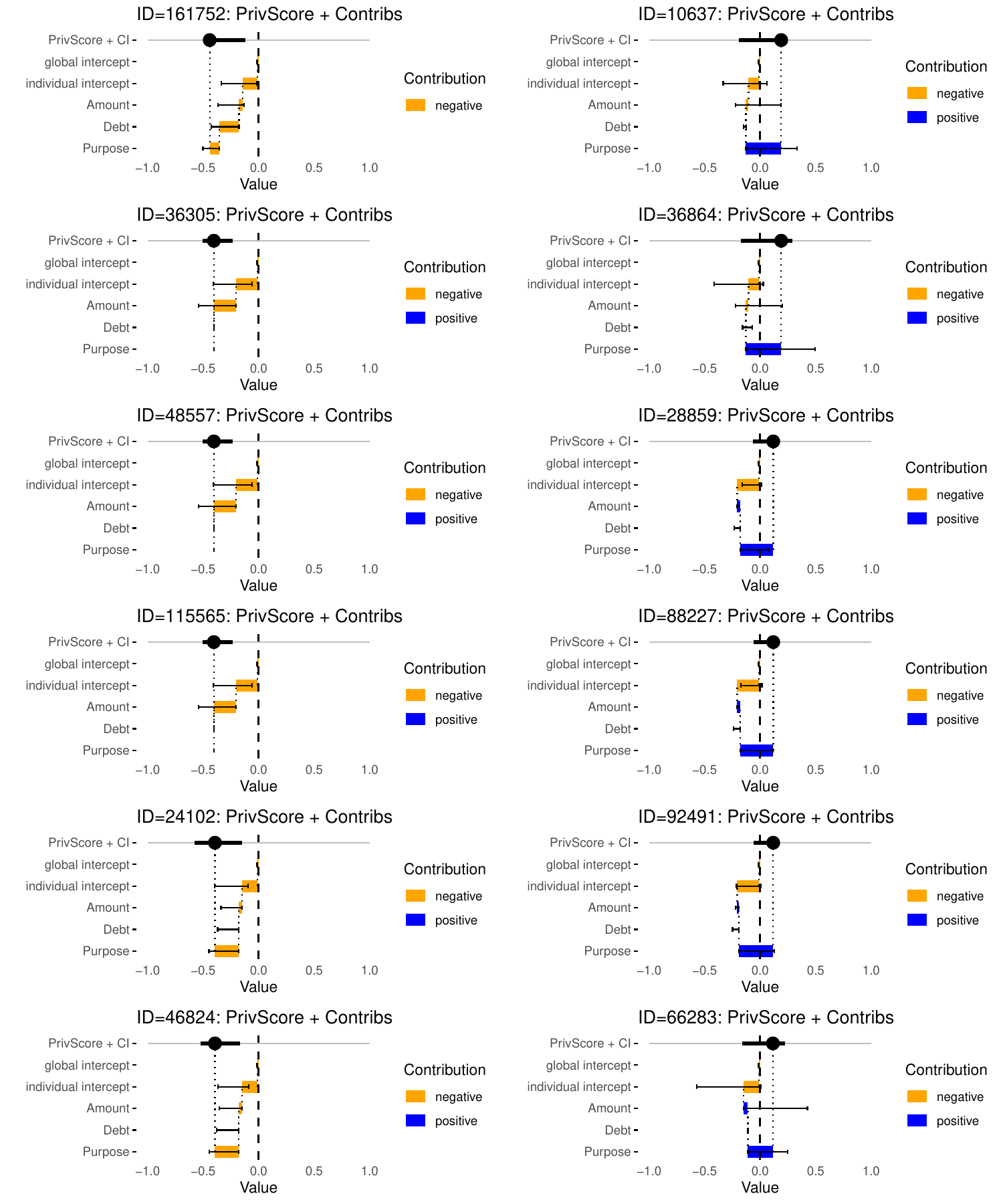} 
    \caption{WI -- fairadapt -- Individuals with smallest (left) and largest (right) PS.}
    \label{fig:app-wi-psc-adapt}
\end{figure}

Table \ref{tab:app_wi-adapt} shows the 6 individuals with the lowest PS and the 6 individuals with the highest PS for test data of WI using fairadapt. Figure \ref{fig:app-wi-psc-adapt} shows PSCs for these individuals. 
Table \ref{tab:lm-wi-adapt} exhibits similar sex and race effects as for res-based warping.
Table \ref{tab:psc-wi-fairadapt} shows the PS/PSC results for fairadapt. Numbers are comparable to results of res-based warping.

\begin{sidewaystable}[ht!]
\centering
\caption{WI -- res-based -- Individuals with smallest (upper half) and largest (lower half) PS. Warped features are indicated by ``\_w''.}
\begin{tabular}{rrrrrrrrrrrr}
  \hline
ID & sex & race & age & purpose & amount & debt & purpose\_w & amount\_w & debt\_w & pred\_real & pred\_warped \\ 
  \hline
54405 &   1 &   0 &   0 &   0 & 5000 &   0 & -0.01 & 5000.00 &   0 & 0.18 & 0.99 \\ 
  152974 &   1 &   0 &   1 &   0 & 5000 &   0 & -0.20 & 15000.00 &   0 & 0.25 & 1.00 \\ 
  102617 &   0 &   0 &   0 &   0 & 5000 &   0 & 0.01 & 17414.82 &   0 & 0.33 & 0.98 \\ 
  104638 &   0 &   0 &   0 &   0 & 5000 &   0 & 0.01 & 17414.82 &   0 & 0.33 & 0.98 \\ 
  114935 &   0 &   0 &   0 &   0 & 5000 &   0 & 0.01 & 17414.82 &   0 & 0.33 & 0.98 \\ 
  154885 &   0 &   0 &   0 &   0 & 5000 &   0 & 0.01 & 17414.82 &   0 & 0.33 & 0.98 \\ 
    \hline
  104914 &   0 &   1 &   0 &   1 & 2805000 &   1 & 1.00 & 2805000.00 &   1 & 0.80 & 0.68 \\ 
  14944 &   0 &   0 &   0 &   1 & 5000 &   1 & 1.01 & 17414.82 &   1 & 0.85 & 0.74 \\ 
  32011 &   0 &   0 &   0 &   1 & 5000 &   1 & 1.01 & 17414.82 &   1 & 0.85 & 0.74 \\ 
  49649 &   0 &   0 &   0 &   1 & 5000 &   1 & 1.01 & 17414.82 &   1 & 0.85 & 0.74 \\ 
  66283 &   0 &   0 &   0 &   1 & 5000 &   1 & 1.01 & 17414.82 &   1 & 0.85 & 0.74 \\ 
  110046 &   0 &   0 &   0 &   1 & 5000 &   1 & 1.01 & 17414.82 &   1 & 0.85 & 0.74 \\ 
   \hline
\end{tabular}
\label{tab:app_wi-res}
\end{sidewaystable}

\begin{sidewaystable}[ht!]
\centering
\caption{WI -- fairadapt -- Individuals with smallest (upper half) and largest (lower half) PS. Warped features are indicated by ``\_w''.}
\begin{tabular}{rrrrrrrrrrrr}
  \hline
ID & sex & race & age & purpose & amount & debt & purpose\_w & amount\_w & debt\_w & pred\_real & pred\_warped \\ 
  \hline
161752 &   0 &   0 &   0 &   1 & 55000 &   0 &   0 & 65000 &   1 & 0.41 & 0.85 \\ 
  36305 &   1 &   0 &   0 &   0 & 5000 &   1 &   0 & 15000 &   1 & 0.40 & 0.81 \\ 
  48557 &   1 &   0 &   0 &   0 & 5000 &   1 &   0 & 15000 &   1 & 0.40 & 0.81 \\ 
  115565 &   1 &   0 &   0 &   0 & 5000 &   1 &   0 & 15000 &   1 & 0.40 & 0.81 \\ 
  24102 &   0 &   0 &   0 &   1 & 35000 &   0 &   0 & 45000 &   0 & 0.33 & 0.72 \\ 
  46824 &   0 &   0 &   0 &   1 & 35000 &   0 &   0 & 45000 &   0 & 0.33 & 0.72 \\ 
    \hline
  10637 &   1 &   0 &   0 &   1 & 5000 &   0 &   0 & 15000 &   0 & 0.78 & 0.59 \\ 
  36864 &   1 &   0 &   0 &   1 & 5000 &   0 &   0 & 15000 &   0 & 0.78 & 0.59 \\ 
  28859 &   0 &   0 &   0 &   1 & 285000 &   0 &   0 & 305000 &   0 & 0.88 & 0.76 \\ 
  88227 &   0 &   0 &   0 &   1 & 285000 &   0 &   0 & 305000 &   0 & 0.88 & 0.76 \\ 
  92491 &   0 &   0 &   0 &   1 & 295000 &   0 &   0 & 305000 &   0 & 0.88 & 0.76 \\ 
  66283 &   0 &   0 &   0 &   1 & 5000 &   1 &   0 & 14840 &   1 & 0.82 & 0.70 \\ 
   \hline
\end{tabular}
\label{tab:app_wi-adapt}
\end{sidewaystable}

\begin{table}[ht]
\centering
\caption{WI -- res-based -- Linear Model Summary} 
\begin{tabular}{rrrrr}
  \hline
 & Estimate & Std. Error & t value & Pr($>$$|$t$|$) \\ 
  \hline
(Intercept) & -0.1287 & 0.0012 & -111.1744 & $<$ 1e-16 \\ 
  sex & 0.0018 & 0.0007 & 2.6610 & 0.0077935 \\ 
  race & 0.1019 & 0.0010 & 100.6884 & $<$ 1e-16 \\ 
  purpose & 0.0208 & 0.0007 & 28.1760 & $<$ 1e-16 \\ 
  amount & 0.0000 & 0.0000 & 12.2606 & $<$ 1e-16 \\ 
  debt & 0.0244 & 0.0007 & 36.8238 & $<$ 1e-16 \\ 
  age & 0.0002 & 0.0009 & 0.1831 & 0.8546931 \\ 
   \hline
\end{tabular}
\label{tab:lm_summary-wi-res}
\end{table}

\begin{table}[ht]
\centering
\caption{WI -- res-based -- Mean and quantiles ($\alpha=0.05$) of PS/PSC and PSC importance on test data} 
\begin{tabular}{lrrr}
  \hline
Feature & Mean & Quantiles $(\alpha, 1-\alpha)$ & Importance \\ 
  \hline
$\hp$ & -0.102 & (-0.446, 0.015) & -- \\ 
   $\priv_g$ & -0.011 & (-0.011, -0.011) & 0.011 \\ 
  $\priv_{\xv}$ & -0.111 & (-0.430, 0.080) & 0.140 \\ 
  $\gx[1]$ & -0.003 & (-0.063, 0.048) & 0.019 \\ 
  $\gx[2]$ & -0.005 & (0.000, 0.000) & 0.005 \\ 
  $\gx[3]$ & 0.028 & (0.000, 0.215) & 0.034 \\ 
   \hline
\end{tabular}
\label{tab:psc-wi-res}
\end{table}


\begin{table}[ht]
\centering
\caption{WI -- fairadapt -- Linear Model Summary} 
\begin{tabular}{rrrrr}
  \hline
 & Estimate & Std. Error & t value & Pr($>$$|$t$|$) \\ 
  \hline
(Intercept) & -0.1186 & 0.0005 & -217.6119 & $<$ 1e-16 \\ 
  sex & 0.0025 & 0.0003 & 7.6500 & 2.0637e-14 \\ 
  race & 0.1088 & 0.0005 & 228.2564 & $<$ 1e-16 \\ 
  purpose & 0.0074 & 0.0003 & 21.2066 & $<$ 1e-16 \\ 
  amount & 0.0000 & 0.0000 & 23.8603 & $<$ 1e-16 \\ 
  debt & 0.0014 & 0.0003 & 4.5590 & 5.1585e-06 \\ 
  age & -0.0003 & 0.0004 & -0.7223 & 0.47009 \\ 
   \hline
\end{tabular}
\label{tab:lm-wi-adapt}
\end{table}

\begin{table}[ht]
\centering
\caption{WI -- fairadapt -- Mean and quantiles ($\alpha=0.05$) of PS/PSC and PSC importance on test data} 
\begin{tabular}{lrrr}
  \hline
Feature & Mean & Quantiles $(\alpha, 1-\alpha)$ & Importance \\ 
  \hline
$\hp$ & -0.109 & (-0.241, 0.012) &-- \\ 
   $\priv_g$ & -0.013 & (-0.013, -0.013) & 0.013 \\ 
  $\priv_{\xv}$ & -0.090 & (-0.168, -0.013) & 0.091 \\ 
  $\gx[1]$ & -0.012 & (-0.069, 0.024) & 0.021 \\ 
  $\gx[2]$ & -0.004 & (0.000, 0.000) & 0.004 \\ 
  $\gx[3]$ & 0.010 & (0.000, 0.123) & 0.013 \\ 
   \hline
\end{tabular}
\label{tab:psc-wi-fairadapt}
\end{table}

\subsubsection{Results New York County}

\begin{figure}[ht]
    \centering
    \includegraphics[width=1\textwidth]{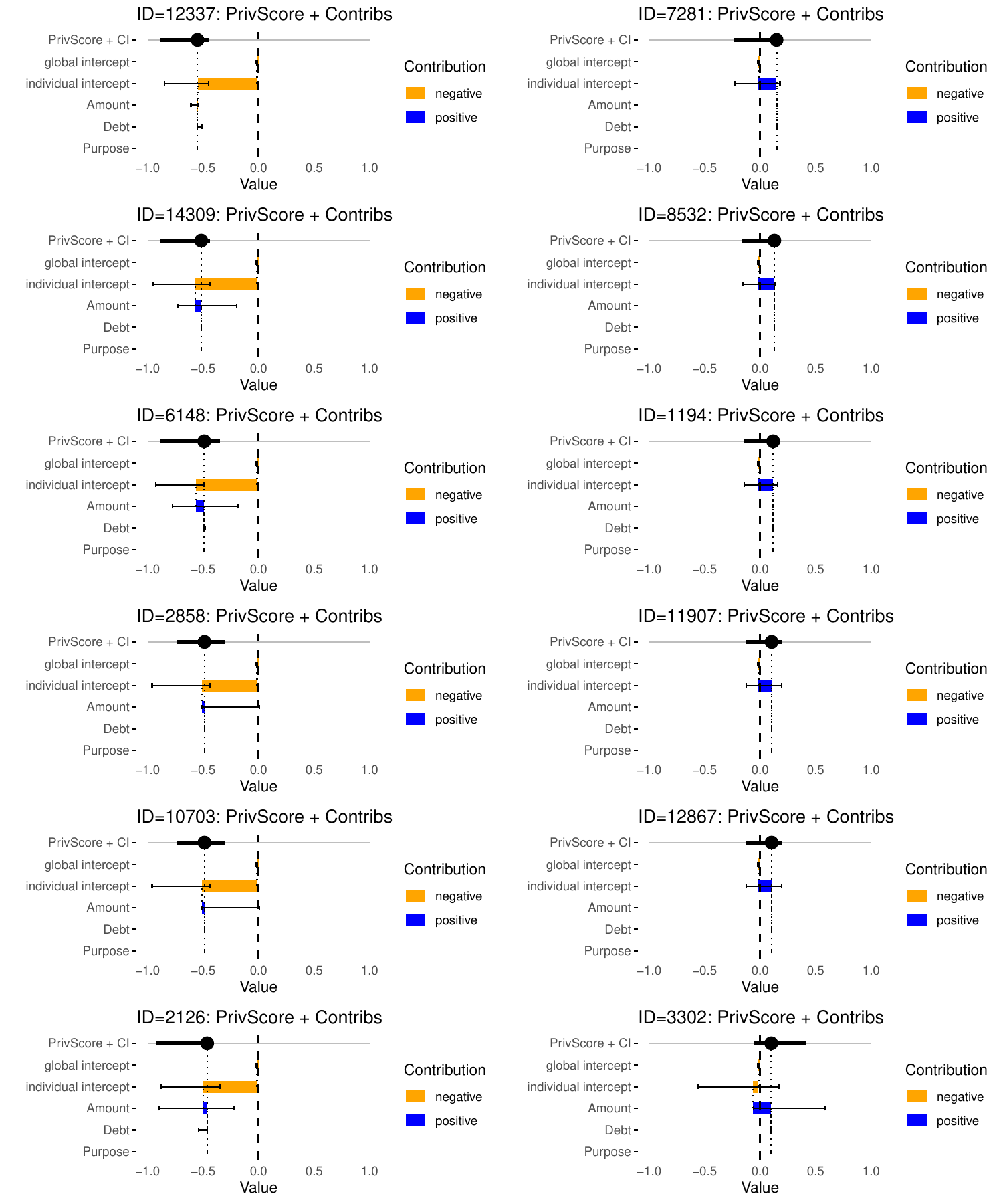} 
    \caption{NY -- res-based -- Individuals with smallest (left) and largest (right) PS.}
    \label{fig:app-ny-psc}
\end{figure}

Table \ref{tab:app_ny-res} shows the 6 individuals with the lowest PS and the 6 individuals with the highest PS for test data of NY using res-based warping. Figure \ref{fig:app-ny-psc} shows PSCs for these individuals.  
Table \ref{tab:lm_summary-ny-res} gives a model summary of regressing PS on real-world features.
Besides the above mentioned race effect of $0.04$ on the PS, we also see a significant -- but smaller and negative -- sex effect of $-0.01$. The age effect is also significantly non-zero. Table \ref{tab:psc-ny-res} shows the PS/PSC results.

\begin{figure}[ht]
    \centering
    \includegraphics[width=1\textwidth]{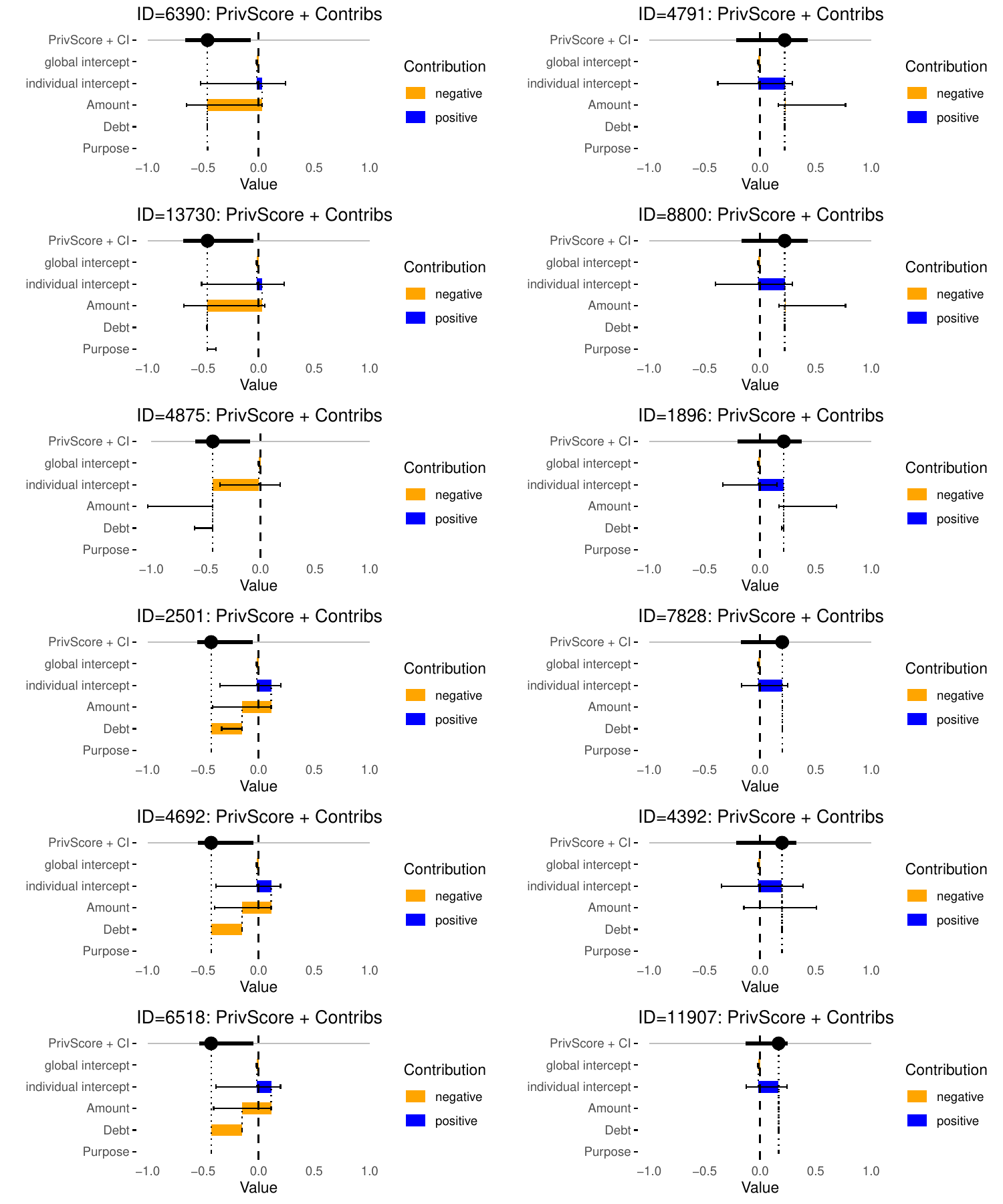} 
    \caption{NY -- fairadapt -- Individuals with smallest (left) and largest (right) PS.}
    \label{fig:app-ny-psc-adapt}
\end{figure}

Table \ref{tab:app_ny-adapt} shows the 6 individuals with the lowest PS and the 6 individuals with the highest PS for test data of NY using fairadapt. Figure \ref{fig:app-ny-psc-adapt} shows PSCs for these individuals. 
Table \ref{tab:lm-ny-adapt} exhibits similar sex, race, and age effects as for res-based warping.
Table \ref{tab:psc-ny-fairadapt} shows the PS/PSC results for fairadapt. Numbers are comparable to results of res-based warping.

\begin{sidewaystable}[ht!]
\centering
\caption{NY -- res-based -- Individuals with smallest (upper half) and largest (lower half) PS. Warped features are indicated by ``\_w''.}
\begin{tabular}{rrrrrrrrrrrr}
  \hline
ID & sex & race & age & purpose & amount & debt & purpose\_w & amount\_w & debt\_w & pred\_real & pred\_warped \\ 
  \hline
12337 &   0 &   0 &   1 &   0 & 15000 &   0 & -0.22 & -84770.46 &   0 & 0.32 & 0.87 \\ 
  14309 &   0 &   0 &   0 &   0 & 45000 &   0 & 0.00 & 28242.03 &   0 & 0.36 & 0.88 \\ 
  6148 &   0 &   0 &   1 &   0 & 105000 &   0 & -0.22 & 14388.93 &   0 & 0.39 & 0.88 \\ 
  2858 &   0 &   0 &   0 &   0 & 55000 &   0 & 0.00 & 38242.03 &   0 & 0.39 & 0.88 \\ 
  10703 &   0 &   0 &   0 &   0 & 55000 &   0 & 0.00 & 38242.03 &   0 & 0.39 & 0.88 \\ 
  2126 &   0 &   0 &   1 &   0 & 145000 &   0 & -0.22 & 61146.90 &   0 & 0.42 & 0.88 \\ 
    \hline
  7281 &   0 &   1 &   0 &   1 & 65000 &   1 & 1.00 & 65000.00 &   1 & 0.74 & 0.60 \\ 
  8532 &   0 &   1 &   0 &   1 & 75000 &   1 & 1.00 & 75000.00 &   1 & 0.77 & 0.64 \\ 
  1194 &   0 &   1 &   0 &   0 & 95000 &   1 & 0.00 & 95000.00 &   1 & 0.55 & 0.43 \\ 
  11907 &   0 &   1 &   0 &   0 & 85000 &   1 & 0.00 & 85000.00 &   1 & 0.53 & 0.43 \\ 
  12867 &   0 &   1 &   0 &   0 & 85000 &   1 & 0.00 & 85000.00 &   1 & 0.53 & 0.43 \\ 
  3302 &   0 &   0 &   1 &   0 & 225000 &   1 & -0.22 & 141146.90 &   1 & 0.75 & 0.65 \\ 
   \hline
\end{tabular}
\label{tab:app_ny-res}
\end{sidewaystable}

\begin{sidewaystable}[ht!]
\centering
\caption{NY -- fairadapt -- Individuals with smallest (upper half) and largest (lower half) PS. Warped features are indicated by ``\_w''.}
\begin{tabular}{rrrrrrrrrrrr}
  \hline
ID & sex & race & age & purpose & amount & debt & purpose\_w & amount\_w & debt\_w & pred\_real & pred\_warped \\ 
  \hline
6390 &   0 &   0 &   0 &   1 & 205000 &   0 &   1 & 214100.00 &   0 & 0.34 & 0.80 \\ 
  13730 &   0 &   0 &   0 &   1 & 205000 &   0 &   1 & 214100.00 &   0 & 0.34 & 0.80 \\ 
  4875 &   1 &   0 &   1 &   0 & 205000 &   0 &   0 & 205000.00 &   0 & 0.26 & 0.70 \\ 
  2501 &   0 &   0 &   0 &   0 & 205000 &   0 &   0 & 214100.00 &   1 & 0.35 & 0.77 \\ 
  4692 &   0 &   0 &   0 &   0 & 205000 &   0 &   0 & 214100.00 &   1 & 0.35 & 0.77 \\ 
  6518 &   0 &   0 &   0 &   0 & 205000 &   0 &   0 & 214100.00 &   1 & 0.35 & 0.77 \\ 
    \hline
  4791 &   1 &   0 &   1 &   0 & 85000 &   0 &   0 & 94740.00 &   0 & 0.64 & 0.42 \\ 
  8800 &   1 &   0 &   1 &   0 & 85000 &   0 &   0 & 94740.00 &   0 & 0.64 & 0.42 \\ 
  1896 &   1 &   0 &   0 &   0 & 95000 &   0 &   0 & 94840.00 &   0 & 0.63 & 0.42 \\ 
  7828 &   0 &   1 &   1 &   0 & 75000 &   1 &   0 & 75000.00 &   1 & 0.61 & 0.42 \\ 
  4392 &   0 &   0 &   0 &   0 & 95000 &   1 &   0 & 94840.00 &   1 & 0.62 & 0.43 \\ 
  11907 &   0 &   1 &   0 &   0 & 85000 &   1 &   0 & 85000.00 &   1 & 0.59 & 0.42 \\ 
   \hline
\end{tabular}
\label{tab:app_ny-adapt}
\end{sidewaystable}

\begin{table}[ht]
\centering
\caption{NY -- res-based -- Linear Model Summary} 
\begin{tabular}{rrrrr}
  \hline
 & Estimate & Std. Error & t value & Pr($>$$|$t$|$) \\ 
  \hline
(Intercept) & -0.0733 & 0.0025 & -29.4392 & $<$ 1e-16 \\ 
  sex & -0.0124 & 0.0018 & -6.8114 & 1.1676e-11 \\ 
  race & 0.0432 & 0.0018 & 23.6297 & $<$ 1e-16 \\ 
  purpose & 0.0515 & 0.0019 & 27.6956 & $<$ 1e-16 \\ 
  amount & -0.0000 & 0.0000 & -0.4767 & 0.63357740 \\ 
  debt & 0.0158 & 0.0018 & 8.8364 & $<$ 1e-16 \\ 
  age & -0.0102 & 0.0028 & -3.7072 & 0.00021344 \\ 
   \hline
\end{tabular}
\label{tab:lm_summary-ny-res}
\end{table}

\begin{table}[ht]
\centering
\caption{NY -- res-based -- Mean and quantiles ($\alpha=0.05$) of PS/PSC and PSC importance on test data} 
\begin{tabular}{lrrr}
  \hline
 & Mean & Quantiles $(\alpha, 1-\alpha)$ & Importance \\ 
  \hline
$\hp$ & -0.039 & (-0.177, 0.019) & -- \\ 
   $\priv_g$ & -0.015 & (-0.015, -0.015) & 0.015 \\ 
  $\priv_{\xv}$ & -0.042 & (-0.131, 0.017) & 0.047 \\ 
  $\gx[1]$ & 0.001 & (-0.016, 0.022) & 0.009 \\ 
  $\gx[2]$ & 0.000 & (0.000, 0.000) & 0.000 \\ 
  $\gx[3]$ & 0.017 & (0.000, 0.094) & 0.017 \\ 
   \hline
\end{tabular}
\label{tab:psc-ny-res}
\end{table}


\begin{table}[ht]
\centering
\caption{NY -- fairadapt -- Linear Model Summary} 
\begin{tabular}{rrrrr}
  \hline
 & Estimate & Std. Error & t value & Pr($>$$|$t$|$) \\ 
  \hline
(Intercept) & -0.0833 & 0.0033 & -24.8914 & $<$ 1e-16 \\ 
  sex & -0.0085 & 0.0024 & -3.4929 & 0.00048489 \\ 
  race & 0.0364 & 0.0025 & 14.8277 & $<$ 1e-16 \\ 
  purpose & 0.0445 & 0.0025 & 17.8217 & $<$ 1e-16 \\ 
  amount & 0.0000 & 0.0000 & 5.1142 & 3.353e-07 \\ 
  debt & 0.0323 & 0.0024 & 13.4780 & $<$ 1e-16 \\ 
  age & -0.0100 & 0.0037 & -2.6911 & 0.00716279 \\  
   \hline
\end{tabular}
\label{tab:lm-ny-adapt}
\end{table}

\begin{table}[ht]
\centering
\caption{NY -- fairadapt -- Mean and quantiles ($\alpha=0.05$) of PS/PSC and PSC importance on test data} 
\begin{tabular}{lrrr}
  \hline
Feature & Mean & Quantiles $(\alpha, 1-\alpha)$ & Importance \\ 
  \hline
$\hp$ & -0.037 & (-0.196, 0.041) & -- \\ 
   $\priv_g$ & -0.015 & (-0.015, -0.015) & 0.015 \\ 
  $\priv_{\xv}$ & -0.009 & (-0.136, 0.056) & 0.047 \\ 
  $\gx[1]$ & -0.010 & (-0.090, 0.031) & 0.025 \\ 
  $\gx[2]$ & -0.006 & (0.000, 0.000) & 0.006 \\ 
  $\gx[3]$ & 0.003 & (0.000, 0.000) & 0.003 \\ 
   \hline
\end{tabular}
\label{tab:psc-ny-fairadapt}
\end{table}
\clearpage

\subsection{Law School admission}
\label{app:sec:exp-real:law}

\subsubsection{Motivating example}
\label{app:sec:uc_aff}

\paragraph{Affirmative Action}

Bell, a female Person of Color, applies to law school and is rejected because her probability of successfully completing law school is predicted to be too low to accept her. Because Bell identifies as a member of a historically discriminated-against subgroup, Bell argues: ``If we lived in a fair world, free of sexism and racism, my educational history would have been different: I would have had access to better schools, a better learning environment, less financial worries, etc., resulting in a higher LSAT (law school admission test) score. To break the wheel of perpetuating this discrimination, I should be allowed to attend law school on a full scholarship because my PS is $-0.2$ with a $99\%$ confidence interval of $(-0.3,-0.13)$.'' As this is below the threshold set by the college, her admission is approved.

%
%

\subsubsection{Data + Setup}


We analyze data on law school acceptance rates in the United States. The data come from a survey conducted by \citet{wightman_lasac_1998} and include information on 22,407 law students, ranging from their undergraduate grade point average (GPA), their Law School Admission Test (LSAT) scores, and their bar exam performance (a binary label indicating whether or not the student ultimately passed the bar)\footnote{We use a dataset version of the survey that is available at \url{https://www.kaggle.com/datasets/danofer/law-school-admissions-bar-passage}.}. Furthermore, the study records information on the students' gender and race which we both utilize as protected attributes in two separate evaluations. The reason for this is that current pre-processing methods are not yet able to account for multiple PAs. Future research should extend these to account for intersectionality. 

More precisely, we analyze the effect of gender and race in two isolated estimation setups. In each of the two setups, we assume the DAG presented in Figure \ref{fig:DAG_lawschool} where $A$ isit{Race}. Since the dataset contains information on various race groups, we recode the variable into a binary indicator where (0) stands for Black and (1) for Non-Black students (we consider the latter as the advantaged group). Similar to the simulation study, we compute confidence intervals using $B=100$ bootstrap samples with replacement. 
As we did not find relevant gender effects, we focus in describing results for PA \textit{Race} in the remainder.


Since our outcome variable (whether or not a student passes the bar examination) is binary, we use a random forest classifier for prediction. We split the data in $80\%$ training and $20\%$ test data, tune hyperparameters with 3-fold CV and random search with 25 evaluations on the training data, and report metrics on the test set. We report summary statistics of the data set in Table \ref{tab:lawschool_sumstats}.


\begin{table}[!h] 
    \centering
    \caption{Summary statistics for Lawschool data}
    \label{tab:lawschool_sumstats}
    \resizebox{\ifdim\width>\linewidth\linewidth\else\width\fi}{!}{
    \begin{tabular}{lccccccccc}
        \toprule
        \multicolumn{1}{c}{ } & \multicolumn{3}{c}{All} & \multicolumn{3}{c}{Non-Black} & \multicolumn{3}{c}{Black} \\
        \cmidrule(l{3pt}r{3pt}){2-4} \cmidrule(l{3pt}r{3pt}){5-7} \cmidrule(l{3pt}r{3pt}){8-10}
        Variable & n & mean & std. dev. & n & mean & std. dev. & n & mean & std. dev.\\
        \midrule
        lsat & 22407 & 36.768 & 5.463 & 21064 & 37.234 & 5.095 & 1343 & 29.459 & 5.830\\
        pass\_bar & 22407 & 0.948 & 0.222 & 21064 & 0.959 & 0.199 & 1343 & 0.778 & 0.416\\
        ugpa & 22407 & 3.215 & 0.404 & 21064 & 3.236 & 0.394 & 1343 & 2.890 & 0.428\\
        \bottomrule
    \end{tabular}}
\end{table}

\subsubsection{Results Law School}

\begin{figure}[ht] 
    \centering
    \includegraphics[width=1\textwidth]{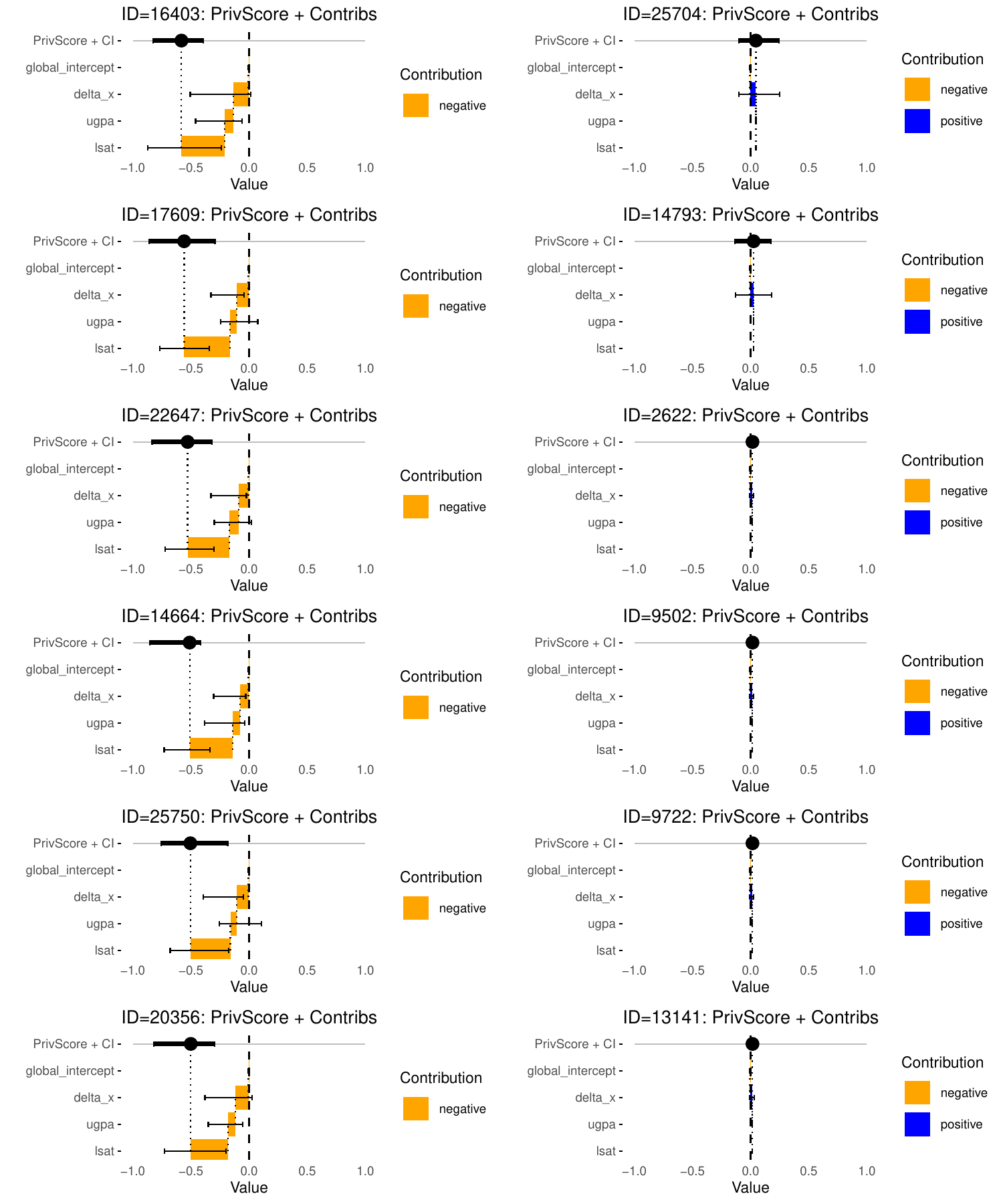} 
    \caption{PS Contribution for the Lawschool data with \textit{Race} as the protected attribute and computed using res-based warping with  bootstrap intervals using $\alpha=0.1$. Individuals with smallest (left) and largest (right) PS.}
    \label{fig:lawschool_resbased_psc}
\end{figure}

Res-based warping: Table \ref{tab:lm_summary-law-res-race} gives a model summary of regressing PS on real-world features.
Besides the a race effect of $0.12$ on the PS, we also see significant -- but smaller -- UGPA and LSAT effects of $0.006$ and $0.003$, respectively. 
Table \ref{tab:psc-law-res-race} shows the PS/PSC results.
Table \ref{tab:lawschool_res-worst} shows the 6 individuals with the lowest PS and the 6 individuals with the highest PS for test data. Figure \ref{fig:lawschool_resbased_psc} shows PSCs for these individuals.  

\begin{figure}[ht] 
    \centering
    \includegraphics[width=1\textwidth]{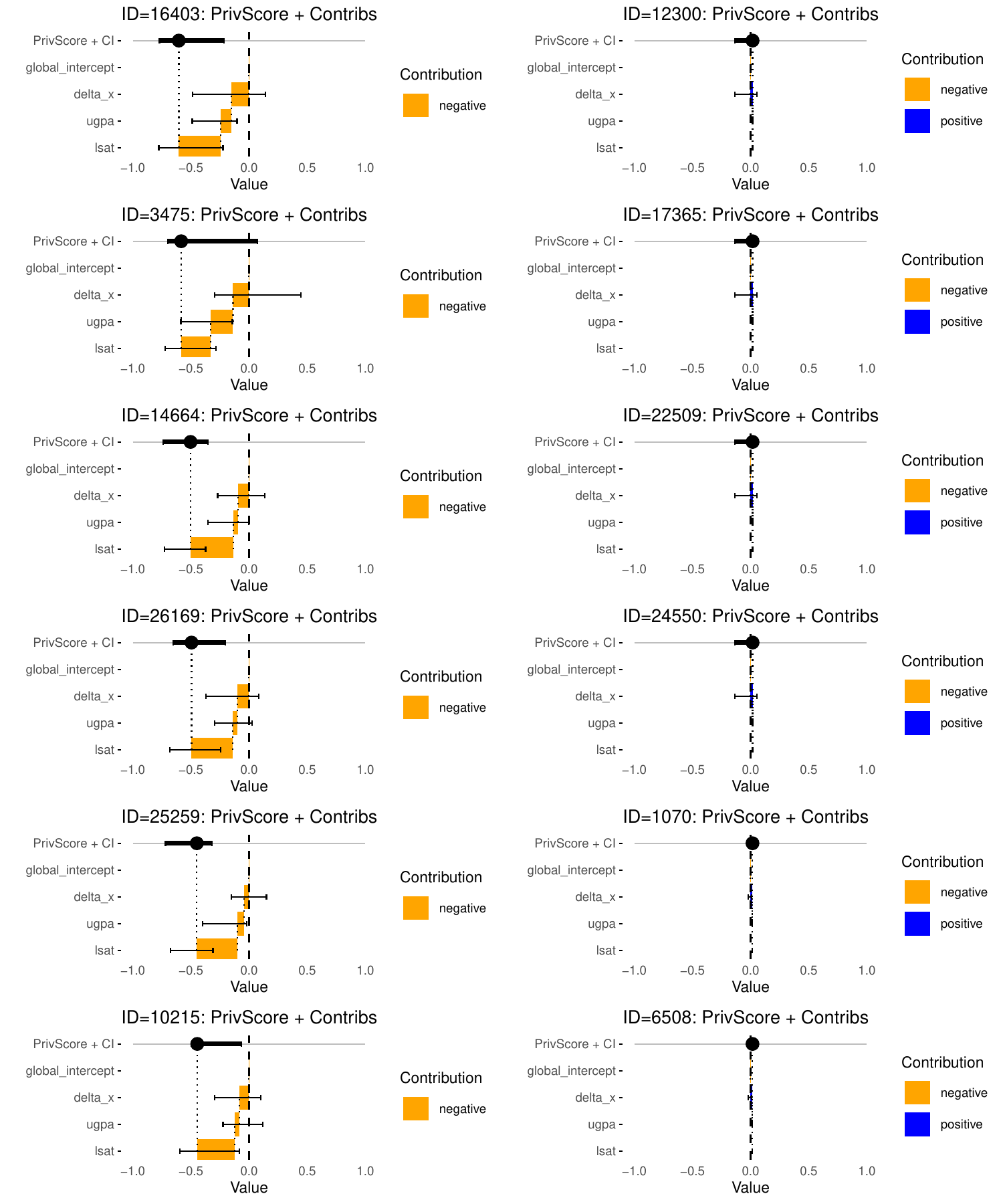} 
    \caption{PS Contribution for the Lawschool data with \textit{Race} as the protected attribute and computed using fairadapt warping with  bootstrap intervals using $\alpha=0.1$. Individuals with smallest (left) and largest (right) PS.}
    \label{fig:lawschool_fairadapt_psc}
\end{figure}

Fairadapt: Table \ref{tab:lm_summary-law-adapt-race} exhibits similar  effects as for res-based warping with the difference that the UGPA effect is not significant.
Table \ref{tab:psc-law-adapt-race} shows the PS/PSC results for fairadapt. Numbers are comparable to results of res-based warping.
Table \ref{tab:lawschool_adapt-worst} shows the 6 individuals with the lowest PS and the 6 individuals with the highest PS for test data. Figure \ref{fig:lawschool_fairadapt_psc} shows PSCs for these individuals. 

\begin{table}[ht] 
    \centering
    \caption{Law School -- res-based -- Linear Model Summary}
    \begin{tabular}{lcccc}
        \hline
        & Estimate & Std. Error & t value & Pr($>$$|$t$|$) \\ 
        \hline
(Intercept) & -0.2679 & 0.0041 & -64.7163 & $<$ 1e-16 \\ 
  race & 0.1198 & 0.0019 & 62.7812 & $<$ 1e-16 \\ 
  ugpa & 0.0064 & 0.0011 & 5.7499 & 9.5245e-09 \\ 
  lsat & 0.0034 & 0.0001 & 39.1490 & $<$ 1e-16 \\ 
   \hline
    \end{tabular} 
    \label{tab:lm_summary-law-res-race}
\end{table}

\begin{table}[ht] 
    \centering
    \caption{Law School -- res-based -- Mean and quantiles ($\alpha=0.05$) of PS/PSC and PSC importance on test data} 
    \begin{tabular}{lccc}
        \hline
        Feature & Mean & Quantiles $(\alpha, 1-\alpha)$ & Importance \\ 
        \hline
$\hp$ & -0.149 & (-0.403, 0.006) & -- \\ 
   $\priv_g$ & -0.005 & (-0.005, -0.005) & 0.005 \\ 
  $\priv_{\xv}$ & -0.010 & (-0.086, 0.018) & 0.026 \\ 
  $\gx[1]$ & -0.022 & (-0.052, 0.002) & 0.023 \\ 
  $\gx[2]$ & -0.111 & (-0.285, -0.001) & 0.111 \\ 
   \hline
    \end{tabular}
    \label{tab:psc-law-res-race}
\end{table}

\begin{table}[ht] 
    \centering
    \caption{Law School -- fairadapt -- Linear Model Summary} 
    \begin{tabular}{lcccc}
        \hline
        & Estimate & Std. Error & t value & Pr($>$$|$t$|$) \\ 
        \hline
(Intercept) & -0.2545 & 0.0040 & -64.2376 & $<$ 1e-16 \\ 
  race & 0.1341 & 0.0018 & 73.4207 & $<$ 1e-16 \\ 
  ugpa & 0.0002 & 0.0011 & 0.1656 & 0.86844 \\ 
  lsat & 0.0032 & 0.0001 & 38.7135 & $<$ 1e-16 \\ 
   \hline
    \end{tabular}
    \label{tab:lm_summary-law-adapt-race}
\end{table}

\begin{table}[ht] 
    \centering
    \caption{Law School -- fairadapt -- Mean and quantiles ($\alpha=0.05$) of PS/PSC and PSC importance on test data} 
    \begin{tabular}{lccc}
        \hline
        Feature & Mean & Quantiles $(\alpha, 1-\alpha)$ & Importance \\ 
        \hline
$\hp$ & -0.159 & (-0.402, -0.020) & -- \\ 
   $\priv_g$ & -0.005 & (-0.005, -0.005) & 0.005 \\ 
  $\priv_{\xv}$ & -0.018 & (-0.068, 0.018) & 0.023 \\ 
  $\gx[1]$ & -0.022 & (-0.049, 0.004) & 0.023 \\ 
  $\gx[2]$ & -0.115 & (-0.292, -0.000) & 0.115 \\ 
   \hline
    \end{tabular}
    \label{tab:psc-law-adapt-race}
\end{table}

\begin{table}[ht!] 
    \centering
        \caption{Lawschool -- res-based -- Individuals with smallest (upper half) and largest (lower half) PS. Warped features are indicated by ``\_w''.}
            \label{tab:lawschool_res-worst}
\begin{tabular}{rrrrrrrrr}
      \hline
    ID & race & ugpa & lsat & pred\_real & ugpa\_w & lsat\_w & pred\_warped \\ 
      \hline
    16403 & 0 & 2.0 & 21.0 & 0.3172 & 2.2 & 29.0 & 0.8990 \\
    17609 & 0 & 2.2 & 20.5 & 0.3408 & 2.5 & 29.0 & 0.8994 \\
    22647 & 0 & 2.4 & 20.3 & 0.3753 & 2.7 & 29.0 & 0.9040 \\
    14664 & 0 & 2.6 & 19.5 & 0.4202 & 3.0 & 27.3 & 0.9316 \\
    25750 & 0 & 2.2 & 20.7 & 0.3946 & 2.5 & 29.0 & 0.8994 \\
    20356 & 0 & 2.5 & 17.0 & 0.3175 & 2.9 & 24.0 & 0.8191 \\
        \hline
    25704 & 1 & 2.5 & 23.0 & 0.6526 & 2.5 & 23.0 & 0.6073 \\
    14793 & 1 & 2.2 & 22.7 & 0.4826 & 2.2 & 22.7 & 0.4568 \\
    2622  & 1 & 2.9 & 46.0 & 0.9786 & 2.9 & 46.0 & 0.9625 \\
    9502  & 1 & 2.9 & 47.0 & 0.9786 & 2.9 & 47.0 & 0.9625 \\
    9722  & 1 & 2.9 & 46.0 & 0.9786 & 2.9 & 46.0 & 0.9625 \\
    13141 & 1 & 2.9 & 48.0 & 0.9786 & 2.9 & 48.0 & 0.9625 \\
    \hline
\end{tabular}
\end{table}

\begin{table}[ht!] 
    \centering
        \caption{Lawschool - fairadapt -- Individuals with smallest (upper half) and largest (lower half) PS. Warped features are indicated by ``\_w''.}
         \label{tab:lawschool_adapt-worst}
\begin{tabular}{rrrrrrrrr}
      \hline
    ID & race & ugpa & lsat & pred\_real & ugpa\_w & lsat\_w & pred\_warped \\ 
      \hline
    16403 & 0 & 2.0 & 21.0 & 0.3367 & 2.1994 & 30.5000 & 0.9413 \\
    3475  & 0 & 2.1 & 15.0 & 0.2058 & 2.4000 & 23.9982 & 0.7899 \\
    14664 & 0 & 2.6 & 19.5 & 0.4328 & 3.0000 & 28.0000 & 0.9364 \\
    26169 & 0 & 2.6 & 18.0 & 0.4340 & 3.0000 & 27.0000 & 0.9292 \\
    25259 & 0 & 3.0 & 20.0 & 0.4648 & 3.4000 & 29.5000 & 0.9145 \\
    10215 & 0 & 2.7 & 18.0 & 0.4775 & 3.2000 & 27.0000 & 0.9236 \\
    \hline
    12300 & 1 & 2.5 & 41.0 & 0.9451 & 2.5000 & 41.0000 & 0.9281 \\
    17365 & 1 & 2.5 & 41.0 & 0.9451 & 2.5000 & 41.0000 & 0.9281 \\
    22509 & 1 & 2.5 & 41.0 & 0.9451 & 2.5000 & 41.0000 & 0.9281 \\
    24550 & 1 & 2.5 & 41.0 & 0.9451 & 2.5000 & 41.0000 & 0.9281 \\
    1070  & 1 & 2.9 & 39.0 & 0.9687 & 2.9000 & 39.0000 & 0.9525 \\
    6508  & 1 & 2.9 & 39.0 & 0.9687 & 2.9000 & 39.0000 & 0.9525 \\
    \hline
\end{tabular}
\end{table}

\end{document}